\pdfoutput=1
\documentclass[11pt]{article}

\usepackage{acl}
\usepackage{marvosym}
\usepackage{times}
\usepackage{latexsym}
\usepackage{graphicx}
\usepackage{booktabs}
\usepackage{tabularx}
\usepackage{subcaption}
\usepackage{amsmath}% http://ctan.org/pkg/amsmath

\newcolumntype{Z}[1]{>{\raggedleft\let\newline\\\arraybackslash\hspace{0pt}}m{#1}}
\newcolumntype{F}[1]{>{\raggedright\let\newline\\\arraybackslash\hspace{0pt}}m{#1}}
\newcolumntype{C}[1]{>{\centering\let\newline\\\arraybackslash\hspace{0pt}}m{#1}}

\usepackage{microtype}

\usepackage{amssymb}
\usepackage{pifont}% http://ctan.org/pkg/pifont
\newcommand{\cmark}{\ding{51}}%
\newcommand{\xmark}{\ding{55}}%
\newcolumntype{L}{>{\centering\arraybackslash}m{3cm}}
\title{
Misinfo Reaction Frames:\\
Reasoning about Readers' Reactions to News Headlines
}
\newcommand\ai{$^\diamondsuit$}
\newcommand\uw{$^\spadesuit$}
\newcommand\ut{$^\clubsuit$}
\newcommand\cmu{$^\heartsuit$}

\usepackage{xcolor}

\author{Saadia Gabriel\uw \space\space\space Skyler Hallinan\uw \space\space\space Maarten Sap\ai\cmu \space\space\space \textbf{Pemi Nguyen}\uw  \space\space\space \\ \space\space\space \textbf{Franziska Roesner}\uw \space\space\space \textbf{Eunsol Choi} \ut \space\space\space \textbf{Yejin Choi}\uw\ai  \\
\uw Paul G. Allen School of Computer Science \& Engineering, University of Washington \\ 
\ut Department of Computer Science, The University of Texas at Austin \\
\ai Allen Institute for Artificial Intelligence \\ 
\cmu Language Technologies Institute, Carnegie Mellon University \\
\small{\texttt{\{skgabrie,hallisky,peming,franzi,yejin\}@cs.washington.edu}},\\\small{\texttt{maartensap@cmu.edu}}, \small{\texttt{eunsol@utexas.edu}}\\
}
\date{}

\begin{document}
\maketitle
\begin{abstract}
Even to a simple and short news headline, readers \emph{react} in a multitude of ways: cognitively (e.g. inferring the writer's intent), emotionally (e.g. feeling distrust), and behaviorally (e.g. sharing the news with their friends). Such reactions are instantaneous and yet complex, as they rely on factors that go beyond interpreting factual content of news.

We propose \textbf{Misinfo Reaction Frames} (MRF), a pragmatic formalism for modeling how readers might react to a news headline. In contrast to categorical schema, our free-text dimensions provide a more nuanced way of understanding intent beyond being benign or malicious. 
We also introduce a Misinfo Reaction Frames corpus, a crowdsourced dataset of reactions to over 25k news headlines focusing on global crises: the Covid-19 pandemic, climate change, and cancer. 

Empirical results confirm that it is indeed possible for neural models to predict the prominent patterns of readers' reactions to previously unseen news headlines. 
% We also find a potentially positive use case of our model; 
Additionally, our user study shows that displaying machine-generated MRF implications alongside news headlines to readers can increase their trust in real news while decreasing their trust in misinformation.
% When we present inferences generated by our best MRF-trained model to people, we find that the machine inferences can increase readers' trust in real news while decreasing their trust in misinformation.
Our work demonstrates the feasibility and importance of pragmatic inferences on news headlines to help enhance AI-guided misinformation detection and mitigation.
\end{abstract}

\section{Introduction} 

\begin{figure}[!t]
    \centering
    \includegraphics[width=.8\linewidth]{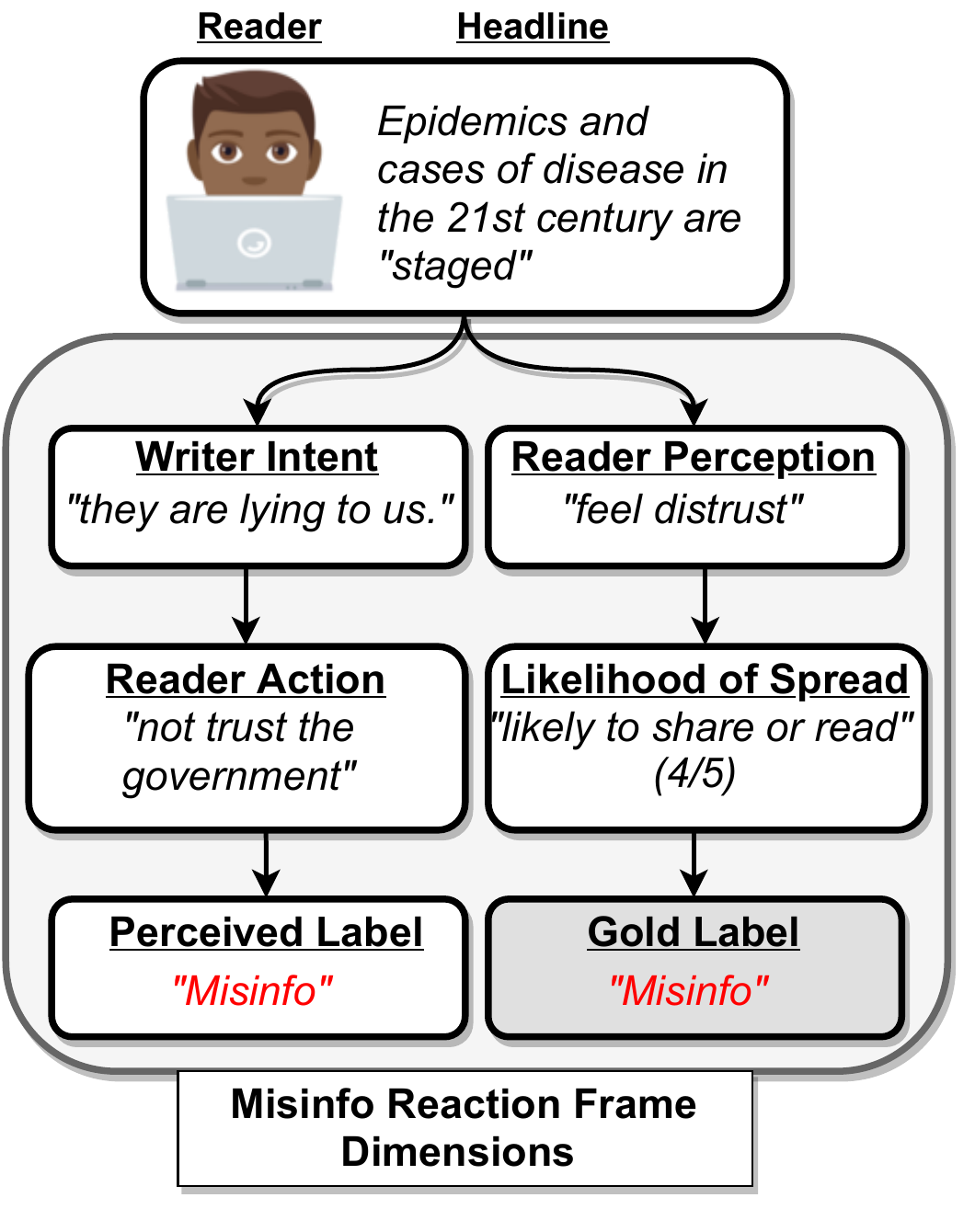}
    \vspace{-1ex}
    \caption{In a binary classification setup, the reaction of the reader is unclear. Here, we use Misinfo Reaction Frames to understand how the reader perceives and reacts to the headline. Our pragmatic frames explain how a health or climate news article is interpreted as reliable or misinformation by readers by incorporating not only linguistic knowledge (e.g. emotions invoked by certain content words), but also knowledge of common social behaviors and domain-specific reasoning. We also include fact-checked labels (gold label).}
    \label{fig:mars}
\end{figure}

\begin{table*}[t]
    \renewcommand{\arraystretch}{0.5}
    \small
    \centering
    \begin{tabular}{F{10em}F{11 em}F{11.5em}cc}
         \PointingHand \textbf{News Headline}   & \WritingHand \textbf{Writer's Intent}  & \WritingHand \textbf{Reader Reaction} & \textbf{Spread} &  \textbf{Real News?}   \\ 
         & & &  & (GPT-2 / T5 / Gold) \\ \midrule
         How COVID is Affecting U.S. Food Supply Chain & \textbf{Human}: ``the pandemic is interrupting the flow of groceries to consumers" \newline \textbf{GPT-2}: ``food supplies are being affected by covid" \newline \textbf{T5}: ``food supply chain is affected by covid" & \textbf{Human}: ``want to know how their groceries will get to them"\newline\textbf{GPT-2}: ``want to learn more" \newline\textbf{T5}:``want to find out more information"  & 4.0 & \Large \color{green} \checkmark \color{black} / \color{green} \checkmark \color{black} / \color{green} \checkmark\\ 
         \\ 
         Thai police arrested a cat for disobeying the curfew order. & \textbf{Human}: ``governments are ludicrous and obtuse."
         \newline \textbf{GPT-2}: ``animals can be dangerous" \newline \textbf{T5}: ``lockdowns are enforced in thailand" & 
         
         \textbf{Human}: ``feel disbelief" \newline \textbf{GPT-2}: ``feel worried"
         \newline \textbf{T5}: ``feel shocked"
         & 1.0 & \Large \color{red} \xmark \color{black} / \color{red} \xmark \color{black} / \color{red} \xmark
         \\
         \\
         Perspective | I'm a black climate expert. Racism derails our efforts to save the planet. &
         \textbf{Human}: ``since climate change will likely affect poorer nations, rich societies are not motivated to help" \newline
         \textbf{GPT-2}: ``racism is bad" \newline \textbf{T5}: ``racism is a problem in society" & \textbf{Human:} ``want to improve their own behavior towards others" \newline \textbf{GPT-2:} ``want to learn more" \newline \textbf{T5}: ``want to take action" & 3.0 & \Large \color{green} \checkmark \color{black} / \color{red} \xmark \color{black} / \color{green} \checkmark\\
        %   $\underbrace{\text{`Happy corals'}}_{\text{loaded language}}\text{: climate crisis sanctuary} \underbrace{\text{teeming with}}_{\text{loaded language}} \text{life found off east Africa}$  & Real \\
        %   & \\
        %   $\underbrace{\text{Pandemic is a fraud}}_{\text{doubt}} \text{and} \underbrace{\text{RT-PCR tests aren't trustworthy}}_{\text{doubt}}$ & Misinfo \\
        %   & \\
        %   $\underbrace{\text{International shipping is killing the climate}}_{\text{appeal to fear}}$ &  Real \\
        %   & \\
        %   $\underbrace{\text{Triple threat of Covid, climate change and conflict}}_{\text{loaded language}} \underbrace{\text{has plunged}}_{\text{loaded language}} \text{millions into need}$ &   Real\\
        %              & \\
        %   $\text{Indonesia's top officials }\underbrace{\text{``dancing without obeying health protocols"}}_{\text{loaded language}} \text{during the pandemic}$ &  Misinfo\\ 
        \bottomrule

    \end{tabular}
    \caption{Example instances in MRF corpus along with generations from reaction inference models fine-tuned on the corpus. We show the {{predicted writer intent}}, {{reader reactions (either a perception or action)}}, and the {{human-annotated likelihood of the headline being shared or read} (Spread)}. The last column (Real News?) shows {{model-predicted and gold label on whether the headline belongs to a real news or misinformation source}}. Our task introduces a new challenge of understanding how news impacts readers. As shown by the examples, large-scale pretrained models (GPT-2, T5) miss nuances present in perceptions of informed readers even when they correctly predict whether the headline is from real news or not.
    %\ms{I wish there was a way to make this table shorter? maybe making the fonts even smaller?}
    } %\ec{i wonder gpt2 and t5 predictions need to be shown here at this point}
    \label{table:examples_data}
\end{table*}

\begin{quote}
\textit{Many objects, persons, and experiences in the world are framed in terms of their potential role in supporting, harming, or enhancing people's lives or interests. We can know that this is so if we know how to interpret expressions in which such things are evaluated...}
\\
- Charles J. Fillmore (1976)
\end{quote}
%\\-Charles J. Fillmore, Frame Semantics and the Nature of Language (1976)

Effectively predicting how a headline may influence a reader requires knowledge of how readers perceive the intent behind real and fake news. 
While most prior NLP research on misinformation has focused on fact-checking, preventing spread of misinformation goes beyond determining veracity \cite{schuster2020limitations,social_motives}.

For example, in Figure \ref{fig:mars}, mistrust in the government may lead readers to share pandemic conspiracy headlines like \textit{``Epidemics and cases of disease in the 21st century are ``staged""} even if they suspect it is misinformation. 
The widespread circulation of misinformation can have serious negative repercussions on readers --- it can reinforce sociopolitical divisions like anti-Asian hate \cite{Vidgen2020DetectingEA,voterfraud}, worsen public health risks \cite{10.1145/3274327}, and undermine efforts to educate the public about global crises \cite{climate}. 

%\ms{This next paragraph should be closer to what we did. Avoid words like ``contextual'' which implies that we're using conversational history/news source? And/or maybe we }
We introduce \textbf{Misinfo Reaction Frames} (MRF), a pragmatic formalism % inspired by Frame semantics \cite{Fillmore1976FRAMESA}, 
to reason about the effect of news headlines on readers. 
Inspired by Frame semantics \cite{Fillmore1976FRAMESA}, our frames distill the pragmatic implications of a news headline in a structured manner.  We capture free-text explanations of readers reactions and perceived author intent, as well as categorical estimates of veracity and likelihood of spread (Table \ref{table:dims}).
% Similarly to semantic frames, our pragmatic frames allow us to consider the meaning of a news headline in a structured manner that incorporates situational commonsense reasoning (e.g. to predict what actions a reader might take as a result). 
%Our free-text dimensions are well-suited for adapting to different news domains and the shifting focus of news events over time. 
We use our new formalism to collect the MRF corpus, a dataset of 202.3k news headline/annotated dimension pairs (69.8k unique implications for 25.1k news headlines) from Covid-19, climate and cancer news. %Furthermore, our pragmatic frames capture richer implications, which rely on situational commonsense reasoning and cognitive psychology that goes beyond what can be modeled with a purely categorical schema. 

We train reaction inference models to predict MRF dimensions from headlines. 
As shown by Table \ref{table:examples_data}, reaction inference models can correctly label the veracity of headlines (85\% F1) and infer commonsense knowledge like \textit{``a cat being arrested for disobeying curfew $\implies$ lockdowns are enforced."} However, models struggle with more nuanced implications \textit{``a cat arrested for disobeying curfew $\implies$ government incompetence."} 
We test generalization of reaction frame inference on a new cancer domain and achieve 86\% F1 by finetuning our MRF model on 574 annotated examples. 

\begin{table*}[t]
    \small
    \centering
    \renewcommand{\arraystretch}{0.5}
    \begin{tabular}{llF{18em}F{12em}}
         \textbf{Dimension}   & \textbf{Type}   & \textbf{Description} & \textbf{Example}  \\ \midrule
         Writer Intent & free-text & A writer intent implication captures \textbf{the readers' interpretation of what the writer is implying.} & ``some masks are better than others."\\
         & \\
         Reader Perception & free-text & A reader perception implication describes how readers would \textit{feel} in response to a headline. These inferences include \textbf{emotional reactions} and \textbf{observations}. & ``feeling angry.", ``feeling that the event described in the headline would trouble most people."\\      
         & \\ 
         Reader Action & free-text & A reader action implication captures what readers would \textit{do} in response to a headline. These describe \textbf{actions}. & ``buy a mask."\\
         & \\ 
         Likelihood of Spread & ordinal & To take into account variability in impact of misinformation due to low or high appeal to readers, we use a 1-5 Likert \cite{Likert1932ATF} scale to measure the \textbf{likelihood of an article being shared or read}. Categories are \{\textit{Very Likely}, \textit{Likely}, \textit{Neutral}, \textit{Unlikely}, \textit{Very Unlikely}\}.  & 4/5 \\
         & \\
         Perceived Label & binary &  We elicit the perceived label (real/misinfo) of a headline, i.e. \textbf{whether it appears to be misinformation or real news to readers}. %This provides information about which headlines are more or less likely to confuse readers. 
         & real\\
         & \\
         Gold Label & binary & We include the \textbf{original ground-truth headline label} (real/misinfo) that was verified by fact-checkers. & misinfo\\
        \bottomrule

    \end{tabular}
    \caption{A description of misinformation reaction frame dimensions. }%\ms{Likelihood of spread is ordinal / Likert scale, right?} 
    \label{table:dims}
\end{table*}

To showcase the usefulness of the MRF framework in user-facing interventions, we investigate the effect of MRF explanations on reader trust in headlines. 
Notably, in a user study our results show that machine-generated MRF inferences affect readers' trust in headlines and for the best model there is a statistically significant correlation (Pearson's $r$=0.24, $p$=0.018) with labels of trustworthiness (\S\ref{sec:gen}). 

Our framework and corpus highlight the need for reasoning about the pragmatic \textit{implications} of news headlines with respect to reader reactions to help combat the spread of misinformation.
We publicly release the MRF corpus and trained models to enable further work (\url{https://github.com/skgabriel/mrf-modeling}).\footnote{The full data annotation setup can be found here: \url{https://misinfo-belief.github.io/}, for use in extending reaction frames to other news domains.} 
We explore promising future directions (and limitations) in (\S\ref{sec:future_work}). 

\section{Misinfo Reaction Frames}
\label{sec:dims}

%\ms{Suggestion: move the relevant related work up here, so we can contrast our approach better while paying hommage to the wealth of existing related work.}
%\ms{As it is now, we need more description of what exactly this formalism is. Peppering in past related work on classification of fake news (to contrast), and then re-explain how it's a formalism with different dimensions, inspired by frame semantics.}
\paragraph{Motivation for Our Formalism} 
In contrast to prior work on misinformation detection \cite{ott-etal-2011-finding, rubin-etal-2016-fake,rashkin-etal-2017-truth,Wang2017LiarLP,Hou2019TowardsAD,volkova-etal-2017-separating,10.1145/3274351} which mostly focuses on linguistic or social media-derived features, we focus on the potential impact of a news headline by modeling readers' reactions.
This approach is to better understand how misinformation can be countered, as it has been shown that interventions from AI agents are better at influencing readers than strangers \cite{Kulkarni2013AllTN}. 

In order to model impact, we build upon prior work that aims to describe the rich interactions involved in human communication, including semantic frames \cite{Fillmore1976FRAMESA}, the encoder-decoder theory of media \cite{Hall1973EncodingAD}\footnote{This theory proposes that before an event is communicated, a narrative discourse encoding the objectives of the writer is generated.}, Grice's conversational maxims \cite{grice1975logic} and the rational speech act model \cite{GOODMAN2016818}\footnote{Here pragmatic interpretation is framed as a probabilistic reasoning problem.}. By describing these interactions with free-text implications invoked by a news headline, we also follow from prior work on pragmatic frames of connotation and social biases \cite{speer-havasi-2012-representing, rashkin-etal-2018-event2mind, Sap2019ATOMICAA, socialbf, socialc}. 

While approaches like rational speech acts model both a pragmatic speaker and listener, we take a \textbf{reader-centric} approach to interpreting ``intent" of a news headline given that the writer's intent is challenging to recover in the dynamic environment of social media news sharing \cite{10.1145/3359229}.
By bridging communication theory, data annotation schema and predictive modeling, we define a concrete framework for understanding the impact of a news headline on a reader.

\paragraph{Defining the Frame Structure} Table \ref{table:examples_data} shows real and misinformation news examples from our dataset with headlines obtained from sources described in \S\ref{sec:data}. We pair these headline examples with generated reaction frame annotations from the MRF corpus. Each reaction frame contains the dimensions in Table \ref{table:dims}.

We elicit annotations based on a \textit{news headline}, which summarizes the main message of an article. We explain this further in \S\ref{sec:data}. An example headline is ``\textit{Covid-19 may strike more cats than believed}." To simplify the task for annotators and ground implications in real-world concerns, we define these implications as relating to one of 7 common themes (e.g. technology or government entities) appearing in Covid and climate news.\footnote{We use a subset of the data (approx. 200 examples per news topic) to manually identify themes. Note that themes are not disjoint and a news article may capture aspects of multiple themes.} We list all the themes in Table \ref{table:themes}, with some themes being shared between topics.

\section{Misinfo Reaction Frames Corpus}

To construct a corpus for studying reader reactions to news headlines, we obtain 69,885 news implications (See \S\ref{sec:data}) by eliciting annotations for 25,164 news headlines (11,757 Covid related articles, 12,733 climate headlines and 674 cancer headlines). There are two stages for collecting the corpus - (1) news data collection and (2) crowd-sourced annotation. 
\subsection{News Data Collection} 
\label{sec:data}
A number of definitions have been proposed for labeling news articles based on reliability. To scope our task, we focus on false news that may be unintentionally spread (misinformation). This differs from disinformation, which assumes a malicious intent or desire to manipulate \cite{Fallis2014AF}. We examine reliable and unreliable headline extracted from two domains with widespread misinformation: Covid-19 \cite{hossain-etal-2020-covidlies} and climate change \cite{lett8fake}. We additionally test on cancer news \cite{10.1145/3394486.3403092} to measure out-of-domain performance. 

\paragraph{Climate Change Dataset}
We retrieve both trustworthy and misinformation headlines related to climate change from NELA-GT-2018-2020 \cite{gruppi2020nelagt2019, norregaard2019nelagt2018}, a dataset of news articles from 519 sources. Each source in this dataset is labeled with a 3-way trustworthy score (reliable / sometimes reliable / unreliable). We discard articles from ``sometimes reliable" sources since the most appropriate label under a binary labeling scheme is unclear. To identify headlines related to climate change, we use keyword filtering.\footnote{We kept any article headline that contained at least one of ``environment," ``climate," ``greenhouse gas," or ``carbon tax." We remove noisy examples obtained using these keywords with manual cleaning.} We also use claims from the SciDCC dataset \cite{mishraneuralnere}, which consists of 11k real news articles from ScienceDaily,\footnote{\url{https://www.sciencedaily.com/}} and Climate-FEVER \cite{diggelmann2021climatefever}, which consists of more than 1,500 true and false climate claims from Wikipedia. We extract claims with either supported or refuted labels in the original dataset.\footnote{The data also includes some claims for which there is not enough info to infer a label. We discard these claims.}  

\begin{table}[]
  \small
    \centering
    \begin{tabular}{l|c|c}
         Theme & Climate & Covid \\ \midrule
         Climate Statistics & \checkmark &  \\
         Natural Disasters & \checkmark& \\
         Entertainment & \checkmark& \\
         Ideology & \checkmark& \\
         Disease Transmission & &\checkmark \\
         Disease Statistics & & \checkmark\\
         Health Treatments & &\checkmark \\
         Protective Gear & &\checkmark\\
         Government Entities & \checkmark & \checkmark \\
         Society &  \checkmark&\checkmark\\
         Technology & \checkmark&\checkmark\\ \bottomrule
    \end{tabular}
    \caption{Themes present in articles by each news topic. Some are covered by both climate and Covid domains, while others are domain specific.  }
    \label{table:themes}
\end{table}

\begin{table}[]
    \centering
    \small
    %\begin{subtable}{\linewidth}
    \begin{tabularx}{\linewidth}{l|r|r|r|r}
         Statistic & Train & Dev. & Test & Cancer\\ \midrule
          Headlines & 19,897 & 2,460 & 2,133 & 674 \\
          Unique Intents & 38,172 & 4,867 &  4,388 & 1,232\\
          Unique Percept. & 2,609 & 538 & 421 & 174 \\
          Unique Actions & 15,036 & 2,176 & 1,739 & 704 \\ 
          Total Pairs & 159,564 & 19,700 & 17,890 & 5,227\\ 
          \bottomrule
    \end{tabularx}

    \caption{Dataset-level breakdown of statistics for MRF corpus.}
    \vspace{1em}
    %\label{table:statsa}
    % \end{subtable}   
    %\begin{subtable}{\linewidth}
    %\begin{tabularx}{1\linewidth}{l|r}
     %    Statistic & Full Data\\ \midrule
     %      Avg. Intents per Event (Climate) & 2.06 \\
     %       Avg. Intents per Event (Covid) & 2.11 \\
     %       Avg. Intents per Event (Cancer) & 1.93 \\
     %      Avg. Perceptions per Event (Climate) & 1.57\\
      %     Avg. Perceptions per Event (Covid) & 1.57\\
    %       Avg. Perceptions per Event (Cancer) & 1.48 \\
     %      Avg. Actions per Event (Climate) & 1.41\\
     %      Avg. Actions per Event (Covid) & 1.49\\
     %      Avg. Actions per Event (Cancer) & 1.43 \\
     %     \bottomrule
    %\end{tabularx}
    %\caption{Topic-level breakdown of statistics for MRF corpus.}%\ec{maybe event instead of headline?}
    
    %\label{table:statsb}
    %     \end{subtable}   
        %\caption{Dataset statistics.}
        \label{table:both_stats}
\end{table}

\paragraph{Covid-19 Dataset}
For trustworthy news regarding Covid-19, we use the CoAID  dataset \cite{cui2020coaid} and a Covid-19 related subset of NELA-GT-2020 \cite{gruppi2020nelagt2019}. CoAID contains 3,565 news headlines from reliable sources. These headlines contain Covid-19 specific keywords and are scraped from nine trustworthy outlets (e.g. the World Health Organization).%\ec{should avoid using abbreviation that's not introduced...} 

For unreliable news (misinformation), we use The CoronaVirusFacts/DatosCoronaVirus Alliance Database, a dataset of over 10,000 mostly false claims related to Covid-19 and the ESOC Covid-19 Misinformation Dataset, which consists of over 200 additional URLs for (mis/dis)information examples.\footnote{\url{https://www.poynter.org}}\footnote{\url{esoc.princeton.edu/publications/esoc-covid-19-misinformation-dataset}} These claims originate from social media posts, manipulated media, and news articles, that have been manually reviewed and summarized by fact-checkers.% Claims can be multiple sentences but are usually the same length as news headlines.\textbf{Check: Are all headlines single sentence?}

\paragraph{Cancer Dataset}

We construct an evaluation set for testing out-of-domain performance using cancer real and misinformation headlines from the DETERRENT dataset \cite{10.1145/3394486.3403092}, consisting of 4.6k real news and 1.4k fake news articles. 

\subsection{Annotation Process} 

In this section we outline the structured annotation interface used to collect the dataset. Statistics for the full dataset are provided in Table \ref{table:both_stats}. 

 %\ec{can we show breakdown b/w health and climate in table 3? what;s the average number of intention per event, perception per event, action per event?} 

\paragraph{Annotation Task Interface}

We use the Amazon Mechanical Turk (MTurk) crowdsourcing platform.\footnote{\url{https://www.mturk.com/}} We provide Figure \ref{fig:marsannotation2} in the Appendix to show the layout of our annotation task. For ease of readability during annotation, we present a headline summarizing the article to annotators, rather than the full text of the article. Annotators then rate veracity and likelihood of spread based on the headline, as well as providing free-text responses for writer intent, reader perception and reader action.\footnote{These news events are either article headlines or claims.} We structure the annotation framework around the themes described in \S\ref{sec:dims}.

\paragraph{Quality Control}
We use a three-stage annotation process for ensuring quality control. In the initial pilot, we select a pool of pre-qualified workers by restricting to workers located in the US who have had at least 99\% of their \textit{human intelligence tasks} (hits) approved and have had at least 5000 hits approved. We approved workers who consistently submitted high-quality annotations for the second stage of our data annotation, in which we assessed the ability of workers to discern between misinformation and real news. We removed workers whose accuracy at predicting the label (real/misinfo) of news headlines fell below 70\%. Our final pool consists of 80 workers who submitted at least three annotations during the pilot tasks.  We achieve pairwise agreement of 79\% on the label predicted by annotators during stage 3, which is comparable to prior work on Covid misinformation \cite{hossain-etal-2020-covidlies}. To account for chance agreement, we also measure Cohen's Kappa $\kappa = .51$, which is considered ``moderate'' agreement. Additional quality control measures were taken as part of our extensive annotation post-processing. For details, see Appendix \ref{sec:postp}.

\paragraph{Annotator Demographics} 

We provided an optional demographic survey to MTurk workers during annotation. Of the 63 annotators who reported ethnicity, 82.54\% identified as White, 9.52\% as Black/African-American, 6.35\% as Asian/Pacific Islander, and 1.59\% as Hispanic/Latino. For self-identified gender, 59\% were male and 41\% were female. Annotators were generally well-educated, with 74\% reporting having a professional degree, college-level degree or higher. Most annotators were between the ages of 25 and 54 (88\%). We also asked annotators for their preferred news sources. New York Times, CNN, Twitter, Washington Post, NPR, Reddit, Reuters, BBC, YouTube and Facebook were reported as the 10 most common news sources.

\section{Modeling Reaction Frames}
We test the ability of large-scale language models to predict Misinfo Reaction Frames. For free-text inferences (e.g. writer intent, reader perception), we use generative language models, specifically T5 encoder-decoder \cite{Raffel2020ExploringTL} and GPT-2 decoder-only models \cite{radford2019language}. For categorical inferences (e.g. the gold label), we use either generative models or BERT-based discriminative models \cite{devlin-etal-2019-bert}. We compare neural models to a simple retrieval baseline (\textbf{BERT-NN}) where we use gold implications aligned with the most similar headline from the training set.\footnote{Similarity is measured between headlines embedded with MiniLM, a distilled transformer model \cite{Wang2020MiniLMDS}. We use the Sentence-BERT package \cite{reimers-2019-sentence-bert}.} % All models are based on a transformer architecture \cite{Vaswani2017AttentionIA}. 

\subsection{Controlled Generation} 

For generative models, we use the following input sequence 
\[x = h_1~...~h_T || s_{d} || s_{t},\] 
where $h$ is a headline of length $T$ tokens, $s_t \in \{$[\textit{covid}]$,$[\textit{climate}]$\}$ is a special topic control token, and $s_d$ is a special dimension control token representing one of six reaction frame dimensions. Here $||$ represents concatenation. The output is a short sequence representing the predicted inference (e.g. ``\textit{to protest}'' for reader action, ``\textit{misinfo}'' for the gold label). For GPT-2 models we also append the gold output inference $y = g_1~...~g_N$ during training, where $N$ is the length of the inference.

\paragraph{Inference} 

We predict each token of the output inference starting from the topic token $s_t$ until the [$eos$] special token is generated. In the case of data with unknown topic labels, this allows us to jointly predict the topic label and output inference. We decode using beam search, since generations by beam search are known to be less diverse but more factually aligned with the context \cite{massarelli-etal-2020-decoding}. 

\subsection{Classification} 

For discriminative models, we use the following input sequence 
\[x = [CLS] h_1~...~h_T [SEP], \] 
where [CLS] and [SEP] are model-specific special tokens. The output is a categorical inference. 

\subsection{Training} 

All our models are optimized using cross-entropy loss, where generally for a sequence of tokens $t$
\[CE(t) = - \frac{1}{|t|}\sum_{i=1}^{|t|} log P_\theta(t_i|t_1,...,t_{i-1}).\]
Here $P_\theta$ is the probability given a particular language model $\theta$. Since GPT-2 does not explicitly distinguish between the input and output (target) sequence during training, we take the loss with respect to the full sequence. For T5 we take the loss with respect only to the output. 

\subsection{Masked Fine-Tuning}

To improve generalization of MRF models, we use an additional masked fine-tuning step. We first train a language model $\theta$ on a set of Covid-19 training examples $D_{covid}$ and climate training examples $D_{climate}$. Then we use the Textrank algorithm \cite{mihalcea-tarau-2004-textrank} to find salient keyphrases in $D_{covid}$ and $D_{climate}$, which we term $k_{covid}$ and $k_{climate}$ respectively. We determine domain-specific keyphrases by looking at the complement of $k_{covid} \cap k_{climate}$
\begin{multline*}
   k'_{covid} =  k_{covid} \setminus k_{covid} \cap k_{climate} \\ 
    k'_{climate} = k_{climate} \setminus k_{covid} \cap k_{climate},
\end{multline*}
and only keep the top 100 keyphrases for each domain. We mask out these keyphrases in the training examples from $D_{covid}$ and $D_{climate}$ by replacing them with a $<mask>$ token. Then we continue training by fine-tuning on the masked examples. A similar approach has been shown to improve generalization and reduce shortcutting of reasoning in models for event detection \cite{Liu2020HowDC}. 
%\ec{$t_i$ not defined?}
%, news topic $t$, dimension $d$ and output gold inference $i$, 

%\paragraph{Masked Fine-tuning for Model Robustness}

%In addition to fine-tuning on Misinfo Reaction Frames, we explore a masked fine-tuning approach for improving generalization of MRF models. We transform existing MRF training examples by replacing the top 100 keyphrases for each training set domain with a $<$mask$>$ token. Domain-specific keyphrases are determined by using the Textrank algorithm \cite{mihalcea-tarau-2004-textrank} to construct graphical representations of Covid and climate-related headlines respectively, then rank saliency of word co-occurences in the two domain corpora.  
\section{Experiments} 
%\ec{there should be a more high level sentence here}
%\subsection{Implementation Details} 
%\ec{this should not be the first subsection. :)}
%\ms{Try using \textbackslash S instead of ``section''}
In this section, we evaluate the effectiveness of our proposed framework at predicting likely reactions, countering misinformation and detecting misinformation. We first describe setup for experiments (\S \ref{sec:setup}), as well as evaluation metrics for classification and generation experiments using our corpus (\S\ref{sec:auto},\S\ref{sec:human}). We also show the performance of large-scale language models on the task of generating reaction frames (\S\ref{sec:gen}) and provide results for classification of news headlines (\S\ref{sec:zero}).  

\subsection{Setup}
\label{sec:setup}

We determine the test split according to date to reduce topical and news event overlap between train and test sets.\footnote{We use news articles from 2021 and the last two months of 2020 for the test set. We ensure there is no exact overlap between data splits.} We use the HuggingFace Transformers library \cite{wolf-etal-2020-transformers}. Hyperparameters are provided in Appendix \ref{sec:hyper}. 
%\ms{potentially, move some more of this to the appendix}

%To minimize overlap between training/validation and test, we first do a 50/50 random split of the data (split A and split B). We then take the bottom 20\% of the data in split B in terms of unigram overlap with split A to be the test set. The remaining 80\% of split B is merged with split A, which we divide into training/validation sets using stratified sampling for balanced representation of news topics. (+dates?)

%We set up the train, validation and test sets using 80/10/10 splits of the dataset. 

%To ensure balanced representation of climate and health news across the splits, we used stratified random sampling. 

\subsection{Evaluation Metrics} 
\label{sec:metrics}

\begin{table*}[!htb]
    \centering
    \small
    \begin{tabular}{ll|cccccc}
        &&\multicolumn{2}{c}{Writer Intent }&\multicolumn{2}{c}{Reader Perception}&\multicolumn{2}{c}{Reader Action}\\
         & Model & BLEU-4 $\uparrow$ & BERTScore $\uparrow$ & BLEU-4 $\uparrow$ & BERTScore $\uparrow$ & BLEU-4  $\uparrow$ & BERTScore $\uparrow$ \\ \midrule
            & BERT-NN  & 31.45 & 86.29 & 35.69 & 91.04  & 45.47 & 84.76   \\
          & T5-base  & 51.48 & 88.03 & 31.98 & 92.87 & 53.55 & 85.27   \\
          dev. & T5-large & 51.30 &  \textbf{88.16} & 32.82  & \textbf{92.94} & 57.29  &  \textbf{85.34}  \\
          & GPT-2 (small) &  \textbf{60.68} & 87.35 & \textbf{37.22}  & 92.21 & 54.20 & 84.83  \\
          & GPT-2 (large) & 54.94 & 87.74 & 32.35 & 92.84 & \textbf{57.84} & 85.00 
          \\  \midrule
           & BERT-NN  & 34.46 & 86.35  &  \textbf{37.09}  & 90.84 & 46.57 & 84.78   \\
           test & T5-base & 50.63  & 87.78 & 32.18 & \textbf{93.32} &  57.37 & 85.60  \\             
           & T5-large & 50.86 & \textbf{87.94} & 32.89  & 93.29 & \textbf{62.10} & \textbf{85.88}   \\          
           & GPT-2 (large) & \textbf{60.51} & 87.73 & 34.18 & 92.51 & 59.57  & 85.53\\\bottomrule        
    \end{tabular}
    \caption{Automatic modeling results (generation task). For this table and the following tables, we highlight the best-performing model(s) in \textbf{bold}.}
    \label{table:auto_gen} %\ms{what makes it baseline? isn't it just ``results''? Also, explain what the bolded things mean?}
\end{table*}

\begin{table*}[!htb]
    \small
    \centering
    \begin{tabular}{l|c|llll|c}
    &&\multicolumn{4}{c|}{Influence on Trust  }&\multicolumn{1}{c}{}\\
         Model   & Quality (1-5)   & +Trust (\%) & -Trust (\%) & Corr w/ Label & Corr w/ Label & Socially Acceptable (\%)   \\ 
        & & & & (all gens) & (quality $\geq$ 3) & \\ \midrule
  T5-base  &  3.61 & 8.33  & 7.82 & \textbf{0.24*} & \textbf{0.30*} & \textbf{75.30}   \\ 
  T5-large   & \textbf{3.74} & 7.73 & 9.76 & -0.03  & 0.09 & 74.66 \\
  GPT-2 (large)  &  3.46   & \textbf{9.70} & \textbf{13.10}  & -0.04  & 0.10 &  74.66 \\
        \bottomrule
    \end{tabular}
    \caption{Human evaluation results (generation task). Cells marked by ``*" are statistically significant for p $<$ .05.}
    %\ms{should we bold things?}}  %\ec{delete ``No" for socially acceptable, as it can be inferred from 100-YES?}}
    \label{table:human_gen}
\end{table*}

We compare reaction inference systems using common automatic metrics. We also use human evaluation to assess quality and potential use of generated writer intent inferences. 

\subsubsection{Automatic Metrics} 
\label{sec:auto}
These metrics include the BLEU (-4) ngram overlap metric \cite{Papineni2002BleuAM} and BERTScore \cite{Zhang2020BERTScoreET}, a model-based metric for measuring semantic similarity between generated inferences and references. For classification we report macro-averaged precision, recall and F1 scores.\footnote{We compute these using scikit-learn: \url{https://scikit-learn.org/stable/index.html}}\footnote{For measuring likelihood of spread, predicted and averaged values are rounded to the nearest integer.}  We use publicly available implementations for all metrics (nltk\footnote{\url{https://www.nltk.org/}} for BLEU). 

\subsubsection{Human Evaluation}
\label{sec:human}
%\ms{potentially, you could condense this section and make it crisper to address: (1) }

For human evaluation, we assess generated inferences using the same pool of qualified workers who annotated the original data. We randomly sample model-generated ``writer's intent" implications from T5 models and GPT-2 large over 196 headlines  where generated implications were unique for each model type.\footnote{98 misinfo and 98 real headlines in the dev. set} We elicit 3 unique judgements per headline. Implications are templated in the form \textit{``The writer is implying that [implication]"} for ease of readability. 

\paragraph{Overall Quality} We ask the annotators to assess the overall quality of generated implications on a 1-5 Likert scale (i.e. whether they are coherent and relevant to the headline without directly copying). 

\paragraph{Influence on Trust} We measure whether generated implications impact readers' perception of news reliability by asking annotators whether a generated implication makes them perceive the news headline as more (+) or less (-) trustworthy. 

\paragraph{Perceived Sociopolitical Acceptability} We ask annotators to rate their perception of the beliefs invoked by an implication in terms of whether they represent a majority (mainstream) or minority  (fringe) viewpoint.\footnote{We refer to ``minority" viewpoint broadly in terms of less frequently adopted or extreme social beliefs, rather than in terms of viewpoints held by historically marginalized groups.}

%ms{Maybe rename this section to "Testing the influence of inference on human judgments" ? And potentially separate out this as a subsubsection (and the other human eval as its own subsubsection)}
\paragraph{A/B Testing} For A/B testing, annotators are initially shown the headline with the generated implication hidden. We ask annotators to rate trustworthiness of headlines on a 1-5 Likert scale, with 1 being clearly misinformation and 5 being clearly real news. After providing this rating, we reveal the generated implication to annotators and have them rate the headline again on the same scale. Annotators were not told whether or not implications were machine-generated, and we advised annotators to mark generated implications that were copies of the headlines as low quality. 

%\begin{table}[!htb]
%    \centering
%    \small
%    \begin{tabularx}{1\linewidth}{l|lll}
%        Model & w/o masked & w/ masked & w/ sup.\\\midrule
%          Prop-BERT & 61.78  & - & - \\
%          BERT-large & 30.07 & - & - \\
%          Covid-BERT & 56.85 & - & - \\
%          GPT-2 (large)  & 43.50 & \textbf{65.99} & 86.99 \\
%          T5-large  & 35.52 & 45.91 & 86.00 \\
 %         \bottomrule        
%    \end{tabularx}
%    \caption{Generalization results on cancer data with and without masked finetuning. Supervised results using a small subset of cancer misinformation are provided under the (w/sup.) column. We report the gold F1 score and bold the best unsupervised model.}
%    \label{tab:outofdomain} %\ms{what makes it baseline? isn't it just ``results''? Also, explain what the bolded things mean?}
%\end{table}

\subsection{Generating Reaction Frames}
\label{sec:gen}
%\ms{Rename to ``Generation results''? (to mirror classification results)? Or change classification results to ``Detecting Misinformation''?}
The automatic evaluation results of our generation task are provided in Table \ref{table:auto_gen}. 

\paragraph{Results} We found that the T5-large model was rated as having slightly higher quality generations than the other model variants (Table \ref{table:human_gen}). Most model generations were rated as being \textit{``socially acceptable"}. However in as many as 25.34\% of judgements, generations were found to be not socially acceptable. 

Interestingly, all models were rated capable of influencing readers to trust or distrust headlines, but effectiveness is dependent on the quality of the generated implication. In particular for T5-base, we found a consistent correlation between the actual label and shifts in trustworthiness scores before and after annotators see the generated writer's intent. Annotators reported that writer intents made real news appear more trustworthy and misinformation less trustworthy.\footnote{While for most models the trend is a decrease in trust for both real news and misinformation, for the T5-base model there is a statistically significant correlation of Pearson's $r=.24$ showing shifts in trust align with gold labels.}  %This is an indicator machine-generated belief frame-based interpretations of headlines may serve as useful aids in countering misinformation. 

% \subsection{Dimension Correlation in Generations} 
% \label{sec:corr_dim}

\subsection{Detecting Misinformation} 
\label{sec:zero}
%\ms{Make more clear what rhetorical techniques are and how they relate to misinfo (potentially give an actual example)}

 \begin{table*}[!htb]
    \centering
    \small
    \begin{tabularx}{1\linewidth}{ll|lll|lll|lll}
        &&\multicolumn{1}{c}{Spread }&\multicolumn{1}{c}{Spread }&\multicolumn{1}{c}{Spread }&\multicolumn{1}{c}{Reader}&\multicolumn{1}{c}{Reader}&\multicolumn{1}{c}{Reader}&\multicolumn{1}{c}{Gold}&\multicolumn{1}{c}{Gold}&\multicolumn{1}{c}{Gold}\\
         & Model & P $\uparrow$ & R $\uparrow$ & F1 $\uparrow$ & P $\uparrow$ & R $\uparrow$ & F1 $\uparrow$ & P $\uparrow$ & R $\uparrow$ & F1 $\uparrow$\\ \midrule
            & Majority Baseline  & 7.11 & 20.00 & 10.49 & 29.61 & 50.00 & 37.20 & 26.32 & 50.00 & 34.49    \\
          & T5-base  & \textbf{29.92} & 27.63 & 22.77 & 81.43 & 76.79 & 77.72 & 87.11 & 87.17 & 87.13   \\
          dev. & T5-large & 29.66 & \textbf{30.08} & \textbf{29.04} & 82.60 & \textbf{78.13} & \textbf{79.04} & 88.21 & 88.06 & 88.12  \\
          & GPT-2 (small) & 26.86 & 23.76 & 22.38 & 78.83 & 77.29 & 77.80 & 84.17 & 83.75 & 83.86    \\
          & GPT-2 (large) & 31.76 & 28.96 & 27.59 & \textbf{82.62} & 77.73 & 78.73 & 90.33 & 88.76 & 89.01  \\
          & Prop-BERT & - & - &  - & - & - & - & 51.82 & 51.09 &  46.43\\
          & BERT-large & - & - &  - & - & -  & - & 89.50 & 89.13 & 89.24\\
          & Covid-BERT & - & - & - & - & -  & - & \textbf{90.79} & \textbf{90.50} &  \textbf{90.60}\\  \midrule 
           & Majority Baseline & 7.78 & 20.00 & 11.20 & 27.00 & 50.00 & 35.07 & 31.41 & 50.00 &  38.58 \\
            & T5-base  & 31.75 & 27.02 & 20.59 & 85.01 & 82.55 & 82.91 & 80.02 & 81.16 & 80.43\\
          test & T5-large & 31.69 & 31.98 & \textbf{30.60} & 86.76 & 84.57 &  \textbf{84.95} & 80.75 & 82.35 & 81.20 \\ 
          & GPT-2 (large) & 34.19 & 27.58 &  18.41 & 83.24 & 83.24 &  82.70 & 80.93 & 82.05 &  81.35\\ 
          & Prop-BERT & - & - &  - & - & - & - & 48.83 & 49.26 & 38.79\\
           & BERT-large & - & - &  - & - & - & - & 79.45 & 81.20 & 79.80 \\
           & Covid-BERT & - & - &  - & - & - & - & \textbf{84.83} & \textbf{86.97} &  \textbf{85.26} \\ \midrule 
           & Prop-BERT & - & - &  - & - & -   &  - & \textbf{72.60} & 65.00 &  61.78 \\
           & BERT-large & - & - &  - & - & -   & - & 23.12 & 43.00 & 30.07 \\ 
           cancer & Covid-BERT & - & - &  - & - & -   & - & 67.87 & 61.00 &  56.85\\
           (unsup.) & GPT-2 (large) & \textbf{27.24} & 23.55 & 10.95 & \textbf{64.38} & 59.21 & 54.43  & 59.16 & 53.00 &  43.50\\
           & T5-large & 21.87 & \textbf{24.95} & 21.12  & 62.08 & \textbf{61.62} & \textbf{61.44}  & 41.13 & 48.00 & 35.52\\
            & GPT-2 (large) + masked & 22.93 & 23.94 & \textbf{21.78}  & 60.06 & 55.69 &  51.00  & 66.03 & \textbf{66.00} &  \textbf{65.99}\\
           & T5-large + masked & 21.38 & 22.79 &  19.57  & 54.84 & 54.41 &  53.66  & 65.26 & 55.00 &  45.91\\ \midrule 
           cancer & GPT-2 (large) + sup & 30.32 & 31.03 &  27.38  & 66.97 & 66.83 & 66.84 & 87.13 & 87.00 & 86.99 \\
           (sup.) & T5-large + sup & 12.17 & 21.67 &  10.51  & 75.30 & 67.95 &  66.15  & 86.00 & 86.00 &  86.00 \\
           \bottomrule        
    \end{tabularx}
    \caption{Automatic modeling results (classification task). For the unsupervised cancer setting (unsup.), all models are trained on covid/climate data only or another news dataset (Prop-BERT). For the supervised setting (sup.), we fine-tune on 574 cancer news examples.}
    \label{table:auto_class} %\ms{what makes it baseline? isn't it just ``results''? Also, explain what the bolded things mean?}
\end{table*}

To test if we can detect misinformation using propagandistic content like \textit{loaded or provocative language} (e.g. ``Covid-19 vaccines may be \textit{the worst threat we face}''), we use a pre-trained BERT propaganda detector \cite{da-san-martino-etal-2019-fine} which we denote here as (\textbf{Prop-BERT}).\footnote{The model predicts if any of 18 known propaganda techniques are used to describe a news event. See the paper for the full list.} For our zero-shot setting, we classify a news event as real if it is not associated with any propaganda techniques and misinformation otherwise. As shown by Table \ref{table:auto_class}, F1 results are considerably lower than task-specific models. This is likely due to the fact both real and misinformation news uses propaganda techniques. % (See Table \ref{table:examples_ours} for examples). 

Neural misinformation detection models are able to outperform humans at identifying misinformation (achieving a max F1 of 85.26 compared to human performance F1 of 75.21\footnote{We count disagreements as being labeled misinformation here, discarding disagreements leads to F1 of 74.97.}), but this is still a nontrivial task for large-scale models. When we use \textbf{Covid-BERT} \cite{Mller2020COVIDTwitterBERTAN}, a variant of BERT pretrained on 160M Covid-related tweets, we see an improvement of 5.46\% over BERT without domain-specific pretraining (Table \ref{table:auto_class}). This indicates greater access to domain-specific data helps in misinformation detection, even if the veracity of claims stated in the data is unknown.

\paragraph{Performance on Out-of-Domain Data} 
\label{sec:out-of-domain}

We test the ability of reaction frames to generalize using 100 cancer-related real and misinformation health news headlines \cite{10.1145/3394486.3403092}, see Table \ref{table:auto_class}. For the misinformation detection task, we evaluate gold F1 using the \textbf{Prop-BERT} zero-shot model, MRF-finetuned BERT-large, \textbf{Covid-BERT}, T5-large and GPT-2 large models. We observe that after one epoch of re-training, masked fine-tuning substantially boosts unsupervised performance of generative MRF models (\textbf{GPT-2 large + masked} and \textbf{T5-large + masked}), making them more robust than BERT variants. We compare this performance against the T5-large and GPT-2 large model finetuned on only 574 cancer examples (\textbf{GPT-2 large + sup} and \textbf{T5-large + sup}), and observe that this leads to a performance increase of up to \textit{43.49}\%, achieving similar F1 performance to our domains with full data supervision.

\section{Future Directions and Limitations of Reaction Frames}
\label{sec:future_work}
Our framework presents new opportunities for studying perceived intent and impact of misinformation, which may also aid in countering and detecting misinformation. 
\paragraph{We can estimate content virality.} 
Given the user-annotated labels for likelihood of reading or sharing, we can estimate whether the information in the associated article is likely to propagate.  
\paragraph{We can analyze the underlying intents behind headlines.} Using annotated writer intents, we can determine common themes and perceived intentions in misinformation headlines across domains (e.g. mistrust of vaccination across medical domains). Given the performance of predictive models highlighted by Tables \ref{table:auto_gen} and \ref{table:human_gen}, we can also extend this analysis to unseen headlines. 
\paragraph{We can categorize headlines by severity of likely outcomes.} False headlines that explicitly incite violence, or otherwise encourage actions that lead to psychological or physical harm (e.g. not vaccinating) may be deemed more malicious than false headlines with more benign consequences (e.g. some examples of satire). Future work may explore categorizing severity of headlines based on potential harms resulting from implications. 
\paragraph{Perceived labels can help us understand which headlines may fool readers.} We can use these labels to determine which types of misinformation headlines appear most like real news to generally knowledgeable readers.  These may also help in designing misinformation countering systems and better adversarial examples to improve robustness of misinformation detection models. 
\paragraph{We can generate counter-narratives to misinformation.}
Our results indicate it is possible to generate effective explanations for the intent of headlines that discourage trust in misinformation (Section \ref{sec:gen}), see Appendix \ref{sec:more_ab_test} for examples. We encourage future work that further improves performance of these models (e.g. through integration of domain knowledge). 

\paragraph{Limitations.} Given these future directions, we also consider key limitations which must be addressed if we move beyond viewing Misinfo Reaction Frames as a proof-of-concept and use the dataset as part of a large-scale system for evaluating or countering misinformation. 

Since we focus on news headlines, the context is limited. The intent of a headline may be different from the actual intent of the corresponding article, especially in the case of clickbait. We find headlines to be suitable as online readers often share headlines without clicking on them \cite{10.1145/2964791.2901462}, however future work may explore extending reaction frames to full news articles. 

There is also annotator and model bias. Readers involved in our data curation and human evaluation studies are ``\textit{generally knowledgeable},'' as proved by their ability to discern misinformation from real news. We see this bias as a potential strength as it allows us to find ways to counter misinformation in cases where readers are well-informed but still believe false information. 
However, annotators may have undesirable political or social biases. 
In such cases, gender bias may lead an annotator to assume that a politician mentioned in a headline is male or to dismiss inequality concerns raised by a scientist belonging to a minority group as ``playing the race card." These biases can also appear in pre-training data, leading to model bias.\footnote{Removing these examples from data curation or trying to control for ``annotator neutrality" does not erase the causes that lead to the existence of these biases. The fact that harmful biases can manifest in the viewpoints of informed readers speaks to the pervasiveness of certain stereotypes.} 
Subjectivity in annotation is a point of discussion in many pragmatic-oriented tasks, e.g. social norm prediction \cite{Jiang2021DelphiTM} and toxicity detection \cite{Halevy2021MitigatingRB,sap2021annotatorsWithAttitudes}. 
We encourage conscious efforts to recruit diverse pools of annotators so multiple perspectives are considered, and future work on modeling reaction frames can consider learning algorithms that mitigate harmful effects of biases, depending on use case \cite{khalifa2021distributional,Gordon2022JuryLI}. 

Lastly, we only consider English-language news and annotate with workers based in the US. It may be that news headlines would be interpreted differently in other languages and cultures.

% Annotations are also subjective, which may limit use to improve misinformation detection models. Subjectivity is a point of discussion in many tasks focused on commonsense reasoning, including social norm \cite{Jiang2021DelphiTM} and toxicity detection \cite{Halevy2021MitigatingRB}.

\section{Conclusion} 

We introduced Misinfo Reaction Frames, a pragmatic formalism for understanding reader perception of news reliability. 
%We use these reaction frames to construct a corpus of news headline implications that explain potential interpretation and reaction of readers. 
We show that machine-generated reaction frames can change perceptions of readers, and while large-scale language models are able to discern between real news and misinformation, there is still room for future work. 
Generated reaction frames can potentially be used in a number of downstream applications, including better understanding of event causality, empathetic response generation and as counter-narratives. 

\section{Ethical Considerations}

There is a risk of frame-based machine-generated reader interpretations being misused to produce more persuasive misinformation. However, understanding the way in which readers perceive and react to news is critical in determining what kinds of misinformation pose the greatest threat and how to counteract its effects. Furthermore, while transformer models have contributed to much of the recent algorithmic progress in NLP research and are the most powerful computational models available to us, work has highlighted shortcomings in their performance on domain-specific text \cite{Moradi2021GPT3MA} and noted that these models can easily detect their own machine-generated misinformation \cite{zellers2019neuralfakenews}. Therefore, we do not see this potential dual-use case as an imminent threat, but urge implementation of systemic changes that would discourage such an outcome in the future - e.g. regulation that would lead to required safety and fairness measures \textit{before} large-scale systems are deployed in the wild \cite{EuropeanCommission}.  
%given the current state of domain-specific transfer using transformer models, 

We emphasize that annotations may reflect \textit{perceptions} and \textit{beliefs} of annotators, rather than universal truths \cite{Britt2019ARA}. Especially considering demographic homogeneity of online crowd-source workers, we urge caution in generalizing beliefs or taking beliefs held in certain social/cultural contexts to be factual knowledge. 
We obtained an Institutional Review Board (IRB) exemption for annotation work, and ensured annotators were fairly paid given time estimations.   

\paragraph{Broader impact.}

The rapid dissemination of information online has led to an increasing problem of falsified or misleading news spread on social media like Twitter, Reddit and Facebook \cite{Vosoughi1146,10.1145/3313831.3376784}. We specifically designed the Misinfo Reaction Frames formalism to allow us to identify and predict \textit{high-impact} misinformation that is more likely to spread. This can allow for future research on factors that make misinformation particularly dangerous, as well as systems that are more effective at mitigating spread. 

%[TODO: write up conclusion]

%Bender and Friedman (2018)

%\ms{I think this section could be a first paragraph with the high level conclusions (only our work), and then one or two paragraphs about the \textit{implications} of our work, including the societal and ethical implications (and then we don't need a specific ethical section either)}

\section*{Acknowledgements}
The authors thank members of the DARPA SemaFor program, UW NLP, the UW CSE 599 social computing class and Amy X. Zhang for helpful discussions, as well as the anonymous reviewers and Akari Asai for comments on the draft. This research is supported in part by NSF (IIS-1714566), NSF (2041894), DARPA MCS program through NIWC Pacific (N66001-19-2-4031), DARPA SemaFor program, and Allen Institute for AI.

\bibliography{anthology,emnlp2020}

\begin{thebibliography}{79}
\expandafter\ifx\csname natexlab\endcsname\relax\def\natexlab#1{#1}\fi

\bibitem[{Abilov et~al.(2021)Abilov, Hua, Matatov, Amir, and
  Naaman}]{voterfraud}
Anton Abilov, Yiqing Hua, Hana Matatov, Ofra Amir, and Mor Naaman. 2021.
\newblock Voterfraud2020: a multi-modal dataset of election fraud claims on
  twitter.
\newblock In \emph{Proceedings of AAAI 2021}.

\bibitem[{Allcott and Gentzkow(2017)}]{10.1257/jep.31.2.211}
Hunt Allcott and Matthew Gentzkow. 2017.
\newblock \href {https://doi.org/10.1257/jep.31.2.211} {Social media and fake
  news in the 2016 election}.
\newblock \emph{Journal of Economic Perspectives}, 31(2):211--36.

\bibitem[{Britt et~al.(2019)Britt, Rouet, Blaum, and Millis}]{Britt2019ARA}
M.~Britt, J.~Rouet, Dylan Blaum, and K.~Millis. 2019.
\newblock A reasoned approach to dealing with fake news.
\newblock \emph{Policy Insights from the Behavioral and Brain Sciences}, 6:101
  -- 94.

\bibitem[{Card et~al.(2015)Card, Boydstun, Gross, Resnik, and
  Smith}]{card-etal-2015-media}
Dallas Card, Amber~E. Boydstun, Justin~H. Gross, Philip Resnik, and Noah~A.
  Smith. 2015.
\newblock \href {https://doi.org/10.3115/v1/P15-2072} {The media frames corpus:
  Annotations of frames across issues}.
\newblock In \emph{Proceedings of the 53rd Annual Meeting of the Association
  for Computational Linguistics and the 7th International Joint Conference on
  Natural Language Processing (Volume 2: Short Papers)}, pages 438--444,
  Beijing, China. Association for Computational Linguistics.

\bibitem[{Cui and Lee(2020)}]{cui2020coaid}
Limeng Cui and Dongwon Lee. 2020.
\newblock Coaid: Covid-19 healthcare misinformation dataset.
\newblock \emph{ArXiv}, abs/2006.00885.

\bibitem[{Cui et~al.(2020)Cui, Seo, Tabar, Ma, Wang, and
  Lee}]{10.1145/3394486.3403092}
Limeng Cui, Haeseung Seo, Maryam Tabar, Fenglong Ma, Suhang Wang, and Dongwon
  Lee. 2020.
\newblock \href {https://doi.org/10.1145/3394486.3403092} {Deterrent: Knowledge
  guided graph attention network for detecting healthcare misinformation}.
\newblock In \emph{Proceedings of the 26th ACM SIGKDD International Conference
  on Knowledge Discovery \& Data Mining}, KDD '20, page 492–502, New York,
  NY, USA. Association for Computing Machinery.

\bibitem[{Da~San~Martino et~al.(2019)Da~San~Martino, Yu, Barr{\'o}n-Cede{\~n}o,
  Petrov, and Nakov}]{da-san-martino-etal-2019-fine}
Giovanni Da~San~Martino, Seunghak Yu, Alberto Barr{\'o}n-Cede{\~n}o, Rostislav
  Petrov, and Preslav Nakov. 2019.
\newblock \href {https://doi.org/10.18653/v1/D19-1565} {Fine-grained analysis
  of propaganda in news article}.
\newblock In \emph{Proceedings of the 2019 Conference on Empirical Methods in
  Natural Language Processing and the 9th International Joint Conference on
  Natural Language Processing (EMNLP-IJCNLP)}, pages 5636--5646, Hong Kong,
  China. Association for Computational Linguistics.

\bibitem[{Devlin et~al.(2019)Devlin, Chang, Lee, and
  Toutanova}]{devlin-etal-2019-bert}
Jacob Devlin, Ming-Wei Chang, Kenton Lee, and Kristina Toutanova. 2019.
\newblock \href {https://doi.org/10.18653/v1/N19-1423} {{BERT}: Pre-training of
  deep bidirectional transformers for language understanding}.
\newblock In \emph{Proceedings of the 2019 Conference of the North {A}merican
  Chapter of the Association for Computational Linguistics: Human Language
  Technologies, Volume 1 (Long and Short Papers)}, pages 4171--4186,
  Minneapolis, Minnesota. Association for Computational Linguistics.

\bibitem[{Dharawat et~al.(2020)Dharawat, Lourentzou, Morales, and
  Zhai}]{dharawat2020drink}
Arkin Dharawat, Ismini Lourentzou, Alex Morales, and ChengXiang Zhai. 2020.
\newblock \href {http://arxiv.org/abs/2010.08743} {Drink bleach or do what now?
  covid-hera: A dataset for risk-informed health decision making in the
  presence of covid19 misinformation}.

\bibitem[{Diggelmann et~al.(2020)Diggelmann, Boyd-Graber, Bulian, Ciaramita,
  and Leippold}]{diggelmann2021climatefever}
Thomas Diggelmann, Jordan Boyd-Graber, Jannis Bulian, Massimiliano Ciaramita,
  and Markus Leippold. 2020.
\newblock Climate-fever: A dataset for verification of real-world climate
  claims.
\newblock In \emph{Tackling Climate Change with Machine Learning Workshop at
  NeurIPS 2020}.

\bibitem[{Ding et~al.(2011)Ding, Maibach, Zhao, Roser-Renouf, and
  Leiserowitz}]{climate}
Ding Ding, Edward~W. Maibach, Xiaoquan Zhao, Connie Roser-Renouf, and Anthony
  Leiserowitz. 2011.
\newblock Support for climate policy and societal action are linked to
  perceptions about scientific agreement.
\newblock \emph{Nature Climate Change}.

\bibitem[{{European Commission}(2021)}]{EuropeanCommission}
{European Commission}. 2021.
\newblock In \emph{Proposal for a regulation of the european parliament and of
  the council laying down harmonised rules on artificial intelligence
  (artificial intelligence act) and amending certain union legislative acts}.

\bibitem[{Fallis(2014)}]{Fallis2014AF}
D.~Fallis. 2014.
\newblock A functional analysis of disinformation.
\newblock In \emph{iConference Proceedings}.

\bibitem[{Field et~al.(2018)Field, Kliger, Wintner, Pan, Jurafsky, and
  Tsvetkov}]{field-etal-2018-framing}
Anjalie Field, Doron Kliger, Shuly Wintner, Jennifer Pan, Dan Jurafsky, and
  Yulia Tsvetkov. 2018.
\newblock \href {https://doi.org/10.18653/v1/D18-1393} {Framing and
  agenda-setting in {R}ussian news: a computational analysis of intricate
  political strategies}.
\newblock In \emph{Proceedings of the 2018 Conference on Empirical Methods in
  Natural Language Processing}, pages 3570--3580, Brussels, Belgium.
  Association for Computational Linguistics.

\bibitem[{Fillmore(1976)}]{Fillmore1976FRAMESA}
C.~J. Fillmore. 1976.
\newblock Frame semantics and the nature of language *.
\newblock \emph{Annals of the New York Academy of Sciences}, 280.

\bibitem[{Forbes et~al.(2020)Forbes, Hwang, Shwartz, Sap, and Choi}]{socialc}
Maxwell Forbes, Jena~D. Hwang, Vered Shwartz, Maarten Sap, and Yejin Choi.
  2020.
\newblock \href {https://doi.org/10.18653/v1/2020.emnlp-main.48} {Social
  chemistry 101: Learning to reason about social and moral norms}.
\newblock In \emph{Proceedings of the 2020 Conference on Empirical Methods in
  Natural Language Processing (EMNLP)}, pages 653--670, Online. Association for
  Computational Linguistics.

\bibitem[{Gabielkov et~al.(2016)Gabielkov, Ramachandran, Chaintreau, and
  Legout}]{10.1145/2964791.2901462}
Maksym Gabielkov, Arthi Ramachandran, Augustin Chaintreau, and Arnaud Legout.
  2016.
\newblock \href {https://doi.org/10.1145/2964791.2901462} {Social clicks: What
  and who gets read on twitter?}
\newblock \emph{SIGMETRICS Perform. Eval. Rev.}, 44(1):179–192.

\bibitem[{Geeng et~al.(2020)Geeng, Yee, and Roesner}]{10.1145/3313831.3376784}
Christine Geeng, Savanna Yee, and Franziska Roesner. 2020.
\newblock \href {https://doi.org/10.1145/3313831.3376784} {Fake news on
  facebook and twitter: Investigating how people (don't) investigate}.
\newblock In \emph{Proceedings of the 2020 CHI Conference on Human Factors in
  Computing Systems}, CHI '20, page 1–14, New York, NY, USA. Association for
  Computing Machinery.

\bibitem[{Ghanem et~al.(2018)Ghanem, Rosso, and
  Rangel}]{ghanem-etal-2018-stance}
Bilal Ghanem, Paolo Rosso, and Francisco Rangel. 2018.
\newblock \href {https://doi.org/10.18653/v1/W18-5510} {Stance detection in
  fake news a combined feature representation}.
\newblock In \emph{Proceedings of the First Workshop on Fact Extraction and
  {VER}ification ({FEVER})}, pages 66--71, Brussels, Belgium. Association for
  Computational Linguistics.

\bibitem[{Ghenai and Mejova(2018)}]{10.1145/3274327}
Amira Ghenai and Yelena Mejova. 2018.
\newblock \href {https://doi.org/10.1145/3274327} {Fake cures: User-centric
  modeling of health misinformation in social media}.
\newblock \emph{Proc. ACM Hum.-Comput. Interact.}, 2(CSCW).

\bibitem[{Golbeck et~al.(2018)Golbeck, Mauriello, Auxier, Bhanushali, Bonk,
  Bouzaghrane, Buntain, Chanduka, Cheakalos, Everett, Falak, Gieringer, Graney,
  Hoffman, Huth, Ma, Jha, Khan, Kori, Lewis, Mirano, Mohn~IV, Mussenden,
  Nelson, Mcwillie, Pant, Shetye, Shrestha, Steinheimer, Subramanian, and
  Visnansky}]{10.1145/3201064.3201100}
Jennifer Golbeck, Matthew Mauriello, Brooke Auxier, Keval~H. Bhanushali,
  Christopher Bonk, Mohamed~Amine Bouzaghrane, Cody Buntain, Riya Chanduka,
  Paul Cheakalos, Jennine~B. Everett, Waleed Falak, Carl Gieringer, Jack
  Graney, Kelly~M. Hoffman, Lindsay Huth, Zhenya Ma, Mayanka Jha, Misbah Khan,
  Varsha Kori, Elo Lewis, George Mirano, William~T. Mohn~IV, Sean Mussenden,
  Tammie~M. Nelson, Sean Mcwillie, Akshat Pant, Priya Shetye, Rusha Shrestha,
  Alexandra Steinheimer, Aditya Subramanian, and Gina Visnansky. 2018.
\newblock \href {https://doi.org/10.1145/3201064.3201100} {Fake news vs satire:
  A dataset and analysis}.
\newblock In \emph{Proceedings of the 10th ACM Conference on Web Science},
  WebSci '18, page 17–21, New York, NY, USA. Association for Computing
  Machinery.

\bibitem[{Goodman and Frank(2016)}]{GOODMAN2016818}
Noah~D. Goodman and Michael~C. Frank. 2016.
\newblock \href {https://doi.org/https://doi.org/10.1016/j.tics.2016.08.005}
  {Pragmatic language interpretation as probabilistic inference}.
\newblock \emph{Trends in Cognitive Sciences}, 20(11):818--829.

\bibitem[{Gordon et~al.(2022)Gordon, Lam, Park, Patel, Hancock, Hashimoto, and
  Bernstein}]{Gordon2022JuryLI}
Mitchell~L. Gordon, Michelle~S. Lam, Joon~Sung Park, Kayur Patel, Jeffrey~T.
  Hancock, Tatsunori Hashimoto, and Michael~S. Bernstein. 2022.
\newblock Jury learning: Integrating dissenting voices into machine learning
  models.
\newblock \emph{CHI}.

\bibitem[{Grice(1975)}]{grice1975logic}
H.~P. Grice. 1975.
\newblock \href {http://www.ucl.ac.uk/ls/studypacks/Grice-Logic.pdf} {Logic and
  conversation}.
\newblock In Peter Cole and Jerry~L. Morgan, editors, \emph{Syntax and
  Semantics: Vol. 3: Speech Acts}, pages 41--58. Academic Press, New York.

\bibitem[{Gruppi et~al.(2020)Gruppi, Horne, and Adalı}]{gruppi2020nelagt2019}
Maurício Gruppi, Benjamin~D. Horne, and Sibel Adalı. 2020.
\newblock \href {http://arxiv.org/abs/2003.08444} {Nela-gt-2019: A large
  multi-labelled news dataset for the study of misinformation in news
  articles}.

\bibitem[{Halevy et~al.(2021)Halevy, Harris, Bruckman, Yang, and
  Howard}]{Halevy2021MitigatingRB}
Matan Halevy, Camille Harris, Amy Bruckman, Diyi Yang, and Ayanna~M. Howard.
  2021.
\newblock Mitigating racial biases in toxic language detection with an
  equity-based ensemble framework.
\newblock \emph{Equity and Access in Algorithms, Mechanisms, and Optimization}.

\bibitem[{Hall(1973)}]{Hall1973EncodingAD}
S.~Hall. 1973.
\newblock \href {https://books.google.com/books?id=j1NJAAAAYAAJ}
  {\emph{Encoding and Decoding in the Television Discourse}}.
\newblock Media series: 1972. Centre for Cultural Studies, University of
  Birmingham.

\bibitem[{Hardalov et~al.(2021)Hardalov, Arora, Nakov, and
  Augenstein}]{Hardalov2021ASO}
Momchil Hardalov, Arnav Arora, Preslav Nakov, and Isabelle Augenstein. 2021.
\newblock A survey on stance detection for mis- and disinformation
  identification.
\newblock \emph{ArXiv}, abs/2103.00242.

\bibitem[{Hossain et~al.(2020)Hossain, Logan~IV, Ugarte, Matsubara, Young, and
  Singh}]{hossain-etal-2020-covidlies}
Tamanna Hossain, Robert~L. Logan~IV, Arjuna Ugarte, Yoshitomo Matsubara, Sean
  Young, and Sameer Singh. 2020.
\newblock \href {https://doi.org/10.18653/v1/2020.nlpcovid19-2.11}
  {{COVIDL}ies: Detecting {COVID}-19 misinformation on social media}.
\newblock In \emph{Proceedings of the 1st Workshop on {NLP} for {COVID}-19
  (Part 2) at {EMNLP} 2020}, Online. Association for Computational Linguistics.

\bibitem[{Hou et~al.(2019)Hou, P{\'e}rez-Rosas, Loeb, and
  Mihalcea}]{Hou2019TowardsAD}
Ruihong Hou, Ver{\'o}nica P{\'e}rez-Rosas, S.~Loeb, and Rada Mihalcea. 2019.
\newblock Towards automatic detection of misinformation in online medical
  videos.
\newblock \emph{ArXiv}, abs/1909.01543.

\bibitem[{Huang and Carley(2020)}]{Huang2020DisinformationAM}
Binxuan Huang and Kathleen~M. Carley. 2020.
\newblock Disinformation and misinformation on twitter during the novel
  coronavirus outbreak.
\newblock \emph{ArXiv}, abs/2006.04278.

\bibitem[{Jahanbakhsh et~al.(2021)Jahanbakhsh, Zhang, Berinsky, Pennycook,
  Rand, and Karger}]{Jahanbakhsh2021ExploringLI}
F.~Jahanbakhsh, Amy~X. Zhang, A.~Berinsky, Gordon Pennycook, David~G. Rand, and
  D.~Karger. 2021.
\newblock Exploring lightweight interventions at posting time to reduce the
  sharing of misinformation on social media.
\newblock \emph{Proceedings of the ACM on Human-Computer Interaction}, 5:1 --
  42.

\bibitem[{Jiang et~al.(2021)Jiang, Hwang, Bhagavatula, Bras, Forbes, Borchardt,
  Liang, Etzioni, Sap, and Choi}]{Jiang2021DelphiTM}
Liwei Jiang, Jena~D. Hwang, Chandrasekhar Bhagavatula, Ronan~Le Bras, Maxwell
  Forbes, Jon Borchardt, Jenny Liang, Oren Etzioni, Maarten Sap, and Yejin
  Choi. 2021.
\newblock Delphi: Towards machine ethics and norms.
\newblock \emph{ArXiv}, abs/2110.07574.

\bibitem[{Jiang and Wilson(2018)}]{10.1145/3274351}
Shan Jiang and Christo Wilson. 2018.
\newblock \href {https://doi.org/10.1145/3274351} {Linguistic signals under
  misinformation and fact-checking: Evidence from user comments on social
  media}.
\newblock \emph{Proc. ACM Hum.-Comput. Interact.}, 2(CSCW).

\bibitem[{Khalifa et~al.(2021)Khalifa, Elsahar, and
  Dymetman}]{khalifa2021distributional}
Muhammad Khalifa, Hady Elsahar, and Marc Dymetman. 2021.
\newblock A distributional approach to controlled text generation.
\newblock \emph{ICLR}.

\bibitem[{Kulkarni and Chi(2013)}]{Kulkarni2013AllTN}
Chinmay Kulkarni and Ed~Chi. 2013.
\newblock All the news that's fit to read: a study of social annotations for
  news reading.
\newblock \emph{Proceedings of the SIGCHI Conference on Human Factors in
  Computing Systems}.

\bibitem[{Lai et~al.(2020)Lai, Liu, and Tan}]{10.1145/3313831.3376873}
Vivian Lai, Han Liu, and Chenhao Tan. 2020.
\newblock \href {https://doi.org/10.1145/3313831.3376873} {"why is 'chicago'
  deceptive?" towards building model-driven tutorials for humans}.
\newblock In \emph{Proceedings of the 2020 CHI Conference on Human Factors in
  Computing Systems}, CHI '20, page 1–13, New York, NY, USA. Association for
  Computing Machinery.

\bibitem[{Lett(2017)}]{lett8fake}
Res Lett. 2017.
\newblock Fake news threatens a climate literate world.
\newblock \emph{Nature Communications}, 8(15460):1.

\bibitem[{Likert(1932)}]{Likert1932ATF}
R.~Likert. 1932.
\newblock A technique for the measurement of attitude scales.
\newblock In \emph{Archives of Psychology, 22 140, 55.}

\bibitem[{Lin(2004)}]{lin-2004-rouge}
Chin-Yew Lin. 2004.
\newblock \href {https://www.aclweb.org/anthology/W04-1013} {{ROUGE}: A package
  for automatic evaluation of summaries}.
\newblock In \emph{Text Summarization Branches Out}, pages 74--81, Barcelona,
  Spain. Association for Computational Linguistics.

\bibitem[{Liu et~al.(2020)Liu, Chen, Liu, Jia, and Sheng}]{Liu2020HowDC}
Jian Liu, Yubo Chen, Kang Liu, Yantao Jia, and Zhicheng Sheng. 2020.
\newblock How does context matter? on the robustness of event detection with
  context-selective mask generalization.
\newblock In \emph{FINDINGS}.

\bibitem[{Loshchilov and Hutter(2019)}]{Loshchilov2019DecoupledWD}
I.~Loshchilov and F.~Hutter. 2019.
\newblock Decoupled weight decay regularization.
\newblock In \emph{ICLR}.

\bibitem[{Massarelli et~al.(2020)Massarelli, Petroni, Piktus, Ott,
  Rockt{\"a}schel, Plachouras, Silvestri, and
  Riedel}]{massarelli-etal-2020-decoding}
Luca Massarelli, Fabio Petroni, Aleksandra Piktus, Myle Ott, Tim
  Rockt{\"a}schel, Vassilis Plachouras, Fabrizio Silvestri, and Sebastian
  Riedel. 2020.
\newblock \href {https://doi.org/10.18653/v1/2020.findings-emnlp.22} {How
  decoding strategies affect the verifiability of generated text}.
\newblock In \emph{Findings of the Association for Computational Linguistics:
  EMNLP 2020}, pages 223--235, Online. Association for Computational
  Linguistics.

\bibitem[{Mihalcea and Tarau(2004)}]{mihalcea-tarau-2004-textrank}
Rada Mihalcea and Paul Tarau. 2004.
\newblock \href {https://www.aclweb.org/anthology/W04-3252} {{T}ext{R}ank:
  Bringing order into text}.
\newblock In \emph{Proceedings of the 2004 Conference on Empirical Methods in
  Natural Language Processing}, pages 404--411, Barcelona, Spain. Association
  for Computational Linguistics.

\bibitem[{Miller(1939)}]{name-calling}
Clyde~R. Miller. 1939.
\newblock The techniques of propaganda.
\newblock \emph{How to Detect and Analyze Propaganda}.

\bibitem[{Mishra and Mittal(2021)}]{mishraneuralnere}
Prakamya Mishra and Rohan Mittal. 2021.
\newblock Neuralnere: Neural named entity relationship extraction for
  end-to-end climate change knowledge graph construction.
\newblock \emph{Tackling Climate Change with Machine Learning Workshop at ICML
  2021}.

\bibitem[{Moradi et~al.(2021)Moradi, Blagec, Haberl, and
  Samwald}]{Moradi2021GPT3MA}
Milad Moradi, Kathrin Blagec, Florian Haberl, and Matthias Samwald. 2021.
\newblock Gpt-3 models are poor few-shot learners in the biomedical domain.
\newblock \emph{ArXiv}, abs/2109.02555.

\bibitem[{Mosleh et~al.(2020)Mosleh, Pennycook, and
  Rand}]{Mosleh2020SelfreportedWT}
Mohsen Mosleh, Gordon Pennycook, and David~G. Rand. 2020.
\newblock Self-reported willingness to share political news articles in online
  surveys correlates with actual sharing on twitter.
\newblock \emph{PLoS ONE}, 15.

\bibitem[{M{\"u}ller et~al.(2020)M{\"u}ller, Salath{\'e}, and
  Kummervold}]{Mller2020COVIDTwitterBERTAN}
M.~M{\"u}ller, M.~Salath{\'e}, and P.~Kummervold. 2020.
\newblock Covid-twitter-bert: A natural language processing model to analyse
  covid-19 content on twitter.
\newblock \emph{ArXiv}, abs/2005.07503.

\bibitem[{Nisbet and Scheufele(2009)}]{https://doi.org/10.3732/ajb.0900041}
Matthew~C. Nisbet and Dietram~A. Scheufele. 2009.
\newblock \href {https://doi.org/https://doi.org/10.3732/ajb.0900041} {What's
  next for science communication? promising directions and lingering
  distractions}.
\newblock \emph{American Journal of Botany}, 96(10):1767--1778.

\bibitem[{Norregaard et~al.(2019)Norregaard, Horne, and
  Adali}]{norregaard2019nelagt2018}
Jeppe Norregaard, Benjamin~D. Horne, and Sibel Adali. 2019.
\newblock \href {http://arxiv.org/abs/1904.01546} {Nela-gt-2018: A large
  multi-labelled news dataset for the study of misinformation in news
  articles}.

\bibitem[{Ott et~al.(2011)Ott, Choi, Cardie, and
  Hancock}]{ott-etal-2011-finding}
Myle Ott, Yejin Choi, Claire Cardie, and Jeffrey~T. Hancock. 2011.
\newblock \href {https://www.aclweb.org/anthology/P11-1032} {Finding deceptive
  opinion spam by any stretch of the imagination}.
\newblock In \emph{Proceedings of the 49th Annual Meeting of the Association
  for Computational Linguistics: Human Language Technologies}, pages 309--319,
  Portland, Oregon, USA. Association for Computational Linguistics.

\bibitem[{Pan et~al.(2018)Pan, Pavlova, Li, Li, Li, and
  Liu}]{contentbasedfakenews}
{Jeff Z.} Pan, Siyana Pavlova, Chenxi Li, Ningxi Li, Yangmei Li, and Jinshuo
  Liu. 2018.
\newblock \href {https://doi.org/10.1007/978-3-030-00671-6_39} {Content based
  fake news detection using knowledge graphs}.
\newblock In \emph{The Semantic Web – ISWC 2018 - 17th International Semantic
  Web Conference, 2018, Proceedings}, Lecture Notes in Computer Science
  (including subseries Lecture Notes in Artificial Intelligence and Lecture
  Notes in Bioinformatics), pages 669--683, Germany. Springer Verlag.
\newblock 17th International Semantic Web Conference, ISWC 2018 ; Conference
  date: 08-10-2018 Through 12-10-2018.

\bibitem[{Papineni et~al.(2002)Papineni, Roukos, Ward, and
  Zhu}]{Papineni2002BleuAM}
Kishore Papineni, S.~Roukos, T.~Ward, and Wei-Jing Zhu. 2002.
\newblock Bleu: a method for automatic evaluation of machine translation.
\newblock In \emph{ACL}.

\bibitem[{Radford et~al.(2019)Radford, Wu, Child, Luan, Amodei, and
  Sutskever}]{radford2019language}
Alec Radford, Jeff Wu, Rewon Child, David Luan, Dario Amodei, and Ilya
  Sutskever. 2019.
\newblock Language models are unsupervised multitask learners.
\newblock \emph{Unpublished manuscript}.

\bibitem[{Raffel et~al.(2020)Raffel, Shazeer, Roberts, Lee, Narang, Matena,
  Zhou, Li, and Liu}]{Raffel2020ExploringTL}
Colin Raffel, Noam~M. Shazeer, Adam Roberts, Katherine Lee, Sharan Narang,
  Michael Matena, Yanqi Zhou, W.~Li, and Peter~J. Liu. 2020.
\newblock Exploring the limits of transfer learning with a unified text-to-text
  transformer.
\newblock \emph{J. Mach. Learn. Res.}, 21:140:1--140:67.

\bibitem[{Rashkin et~al.(2017)Rashkin, Choi, Jang, Volkova, and
  Choi}]{rashkin-etal-2017-truth}
Hannah Rashkin, Eunsol Choi, Jin~Yea Jang, Svitlana Volkova, and Yejin Choi.
  2017.
\newblock \href {https://doi.org/10.18653/v1/D17-1317} {Truth of varying
  shades: Analyzing language in fake news and political fact-checking}.
\newblock In \emph{Proceedings of the 2017 Conference on Empirical Methods in
  Natural Language Processing}, pages 2931--2937, Copenhagen, Denmark.
  Association for Computational Linguistics.

\bibitem[{Rashkin et~al.(2018)Rashkin, Sap, Allaway, Smith, and
  Choi}]{rashkin-etal-2018-event2mind}
Hannah Rashkin, Maarten Sap, Emily Allaway, Noah~A. Smith, and Yejin Choi.
  2018.
\newblock \href {https://doi.org/10.18653/v1/P18-1043} {{E}vent2{M}ind:
  Commonsense inference on events, intents, and reactions}.
\newblock In \emph{Proceedings of the 56th Annual Meeting of the Association
  for Computational Linguistics (Volume 1: Long Papers)}, pages 463--473,
  Melbourne, Australia. Association for Computational Linguistics.

\bibitem[{Reimers and Gurevych(2019)}]{reimers-2019-sentence-bert}
Nils Reimers and Iryna Gurevych. 2019.
\newblock \href {https://arxiv.org/abs/1908.10084} {Sentence-bert: Sentence
  embeddings using siamese bert-networks}.
\newblock In \emph{Proceedings of the 2019 Conference on Empirical Methods in
  Natural Language Processing}. Association for Computational Linguistics.

\bibitem[{Ren et~al.(2021)Ren, Dimant, and Schweitzer}]{social_motives}
Zhiying~(Bella) Ren, Eugen Dimant, and Maurice~E. Schweitzer. 2021.
\newblock \href {http://dx.doi.org/10.2139/ssrn.3919364} {Social motives for
  sharing conspiracy theories}.
\newblock \emph{SSRN}.

\bibitem[{Rubin et~al.(2016)Rubin, Conroy, Chen, and
  Cornwell}]{rubin-etal-2016-fake}
Victoria Rubin, Niall Conroy, Yimin Chen, and Sarah Cornwell. 2016.
\newblock \href {https://doi.org/10.18653/v1/W16-0802} {Fake news or truth?
  using satirical cues to detect potentially misleading news}.
\newblock In \emph{Proceedings of the Second Workshop on Computational
  Approaches to Deception Detection}, pages 7--17, San Diego, California.
  Association for Computational Linguistics.

\bibitem[{Sap et~al.(2019)Sap, Bras, Allaway, Bhagavatula, Lourie, Rashkin,
  Roof, Smith, and Choi}]{Sap2019ATOMICAA}
Maarten Sap, Ronan~Le Bras, Emily Allaway, Chandra Bhagavatula, Nicholas
  Lourie, Hannah Rashkin, Brendan Roof, Noah~A. Smith, and Yejin Choi. 2019.
\newblock Atomic: An atlas of machine commonsense for if-then reasoning.
\newblock In \emph{AAAI}.

\bibitem[{Sap et~al.(2020)Sap, Gabriel, Qin, Jurafsky, Smith, and
  Choi}]{socialbf}
Maarten Sap, Saadia Gabriel, Lianhui Qin, Dan Jurafsky, Noah~A. Smith, and
  Yejin Choi. 2020.
\newblock Social bias frames: Reasoning about social and power implications of
  language.
\newblock In \emph{ACL}.

\bibitem[{Sap et~al.(2021)Sap, Swayamdipta, Vianna, Zhou, Choi, and
  Smith}]{sap2021annotatorsWithAttitudes}
Maarten Sap, Swabha Swayamdipta, Laura Vianna, Xuhui Zhou, Yejin Choi, and
  Noah~A. Smith. 2021.
\newblock \href {https://arxiv.org/abs/2111.07997} {Annotators with attitudes:
  How annotator beliefs and identities bias toxic language detection}.

\bibitem[{Schuster et~al.(2020)Schuster, Schuster, Shah, and
  Barzilay}]{schuster2020limitations}
Tal Schuster, R.~Schuster, Darsh~J. Shah, and Regina Barzilay. 2020.
\newblock The limitations of stylometry for detecting machine-generated fake
  news.
\newblock \emph{Computational Linguistics}.

\bibitem[{Shaver et~al.(1987)Shaver, Schwartz, Kirson, and
  O'Connor}]{Shaver1987EmotionKF}
Phillip~R. Shaver, Judith~C. Schwartz, Donald Kirson, and Cary O'Connor. 1987.
\newblock Emotion knowledge: further exploration of a prototype approach.
\newblock \emph{Journal of personality and social psychology}, 52 6:1061--86.

\bibitem[{Speer and Havasi(2012)}]{speer-havasi-2012-representing}
Robyn Speer and Catherine Havasi. 2012.
\newblock \href
  {http://www.lrec-conf.org/proceedings/lrec2012/pdf/1072_Paper.pdf}
  {Representing general relational knowledge in {C}oncept{N}et 5}.
\newblock In \emph{Proceedings of the Eighth International Conference on
  Language Resources and Evaluation ({LREC}-2012)}, pages 3679--3686, Istanbul,
  Turkey. European Languages Resources Association (ELRA).

\bibitem[{Starbird et~al.(2019)Starbird, Arif, and Wilson}]{10.1145/3359229}
Kate Starbird, Ahmer Arif, and Tom Wilson. 2019.
\newblock \href {https://doi.org/10.1145/3359229} {Disinformation as
  collaborative work: Surfacing the participatory nature of strategic
  information operations}.
\newblock \emph{Proc. ACM Hum.-Comput. Interact.}, 3(CSCW).

\bibitem[{Vidgen et~al.(2020)Vidgen, Botelho, Broniatowski, Guest, Hall,
  Margetts, Tromble, Waseem, and Hale}]{Vidgen2020DetectingEA}
Bertie Vidgen, Austin Botelho, David~A. Broniatowski, E.~Guest, M.~Hall,
  H.~Margetts, Rebekah Tromble, Zeerak Waseem, and Scott~A. Hale. 2020.
\newblock Detecting east asian prejudice on social media.
\newblock \emph{ArXiv}, abs/2005.03909.

\bibitem[{Volkova et~al.(2017)Volkova, Shaffer, Jang, and
  Hodas}]{volkova-etal-2017-separating}
Svitlana Volkova, Kyle Shaffer, Jin~Yea Jang, and Nathan Hodas. 2017.
\newblock \href {https://doi.org/10.18653/v1/P17-2102} {Separating facts from
  fiction: Linguistic models to classify suspicious and trusted news posts on
  twitter}.
\newblock In \emph{Proceedings of the 55th Annual Meeting of the Association
  for Computational Linguistics (Volume 2: Short Papers)}, pages 647--653,
  Vancouver, Canada. Association for Computational Linguistics.

\bibitem[{Vosoughi et~al.(2018)Vosoughi, Roy, and Aral}]{Vosoughi1146}
Soroush Vosoughi, Deb Roy, and Sinan Aral. 2018.
\newblock \href {https://doi.org/10.1126/science.aap9559} {The spread of true
  and false news online}.
\newblock \emph{Science}, 359(6380):1146--1151.

\bibitem[{Wang et~al.(2020)Wang, Wei, Dong, Bao, Yang, and
  Zhou}]{Wang2020MiniLMDS}
Wenhui Wang, Furu Wei, Li~Dong, Hangbo Bao, Nan Yang, and Ming Zhou. 2020.
\newblock Minilm: Deep self-attention distillation for task-agnostic
  compression of pre-trained transformers.
\newblock \emph{ArXiv}, abs/2002.10957.

\bibitem[{Wang(2017)}]{Wang2017LiarLP}
William~Yang Wang. 2017.
\newblock ``liar, liar pants on fire": A new benchmark dataset for fake news
  detection.
\newblock In \emph{ACL}.

\bibitem[{Wolf et~al.(2020)Wolf, Debut, Sanh, Chaumond, Delangue, Moi, Cistac,
  Rault, Louf, Funtowicz, Davison, Shleifer, von Platen, Ma, Jernite, Plu, Xu,
  Le~Scao, Gugger, Drame, Lhoest, and Rush}]{wolf-etal-2020-transformers}
Thomas Wolf, Lysandre Debut, Victor Sanh, Julien Chaumond, Clement Delangue,
  Anthony Moi, Pierric Cistac, Tim Rault, Remi Louf, Morgan Funtowicz, Joe
  Davison, Sam Shleifer, Patrick von Platen, Clara Ma, Yacine Jernite, Julien
  Plu, Canwen Xu, Teven Le~Scao, Sylvain Gugger, Mariama Drame, Quentin Lhoest,
  and Alexander Rush. 2020.
\newblock \href {https://doi.org/10.18653/v1/2020.emnlp-demos.6} {Transformers:
  State-of-the-art natural language processing}.
\newblock In \emph{Proceedings of the 2020 Conference on Empirical Methods in
  Natural Language Processing: System Demonstrations}, pages 38--45, Online.
  Association for Computational Linguistics.

\bibitem[{Yang et~al.(2019)Yang, Niven, and Kao}]{Yang2019FakeND}
Kai-Chou Yang, Timothy Niven, and Hung-Yu Kao. 2019.
\newblock Fake news detection as natural language inference.
\newblock \emph{ArXiv}, abs/1907.07347.

\bibitem[{Yaqub et~al.(2020)Yaqub, Kakhidze, Brockman, Memon, and
  Patil}]{10.1145/3313831.3376213}
Waheeb Yaqub, Otari Kakhidze, Morgan~L. Brockman, Nasir Memon, and Sameer
  Patil. 2020.
\newblock \href {https://doi.org/10.1145/3313831.3376213} {Effects of
  credibility indicators on social media news sharing intent}.
\newblock In \emph{Proceedings of the 2020 CHI Conference on Human Factors in
  Computing Systems}, CHI '20, page 1–14, New York, NY, USA. Association for
  Computing Machinery.

\bibitem[{Zellers et~al.(2019)Zellers, Holtzman, Rashkin, Bisk, Farhadi,
  Roesner, and Choi}]{zellers2019neuralfakenews}
Rowan Zellers, Ari Holtzman, Hannah Rashkin, Yonatan Bisk, Ali Farhadi,
  Franziska Roesner, and Yejin Choi. 2019.
\newblock \href
  {http://papers.nips.cc/paper/9106-defending-against-neural-fake-news.pdf}
  {Defending against neural fake news}.
\newblock In H.~Wallach, H.~Larochelle, A.~Beygelzimer, F.~d'~Alch\'{e}-Buc,
  E.~Fox, and R.~Garnett, editors, \emph{Advances in Neural Information
  Processing Systems 32}, pages 9054--9065. Curran Associates, Inc.

\bibitem[{Zhang et~al.(2020)Zhang, Kishore, Wu, Weinberger, and
  Artzi}]{Zhang2020BERTScoreET}
Tianyi Zhang, V.~Kishore, Felix Wu, Kilian~Q. Weinberger, and Yoav Artzi. 2020.
\newblock Bertscore: Evaluating text generation with bert.
\newblock In \emph{ICLR}.

\bibitem[{Zhou and Zafarani(2020)}]{10.1145/3395046}
Xinyi Zhou and Reza Zafarani. 2020.
\newblock \href {https://doi.org/10.1145/3395046} {A survey of fake news:
  Fundamental theories, detection methods, and opportunities}.
\newblock \emph{ACM Comput. Surv.}, 53(5).

\end{thebibliography}
\bibliographystyle{acl_natbib}
\clearpage
\appendix
\section{}
\label{sec:appendix}
\begin{table*}[t]
    \small
    \centering
    \begin{tabular}{C{5em}C{5em}C{9em}C{9em}}
         \textbf{Likelihood of Spread}  & \textbf{Perceived Label}   & \textbf{Potential Perceptions} & \textbf{Potential Actions}  \\ \midrule
         $<$3 & Misinfo  & `feel lied to',`feel disinterested',`feel disbelief',`feel this is false',`feel suspicious' &  `fact-check this article',`skip this article',`check the facts',`avoid sharing this article',`do something else'\\ \midrule
         \\
        $<$3 & Real or Disagree
         & `feel unsure',`feel like they need more information to process this' & `move on to the next thing',`read more', `learn more'\\ \midrule
         \\
         $>$3 & Any &  `feel curious',`feel interested',`feel like this is something others might want to know about' & `talk to a friend about it',`share the article',`learn more',`read more',`try to understand'\\ \midrule
         \\
         $=$3 & Any & `feel indifferent' & `move on to something else' \\
        \bottomrule

    \end{tabular}
    \caption{Process for handling missing reader annotations.}
    \label{table:reassign}
\end{table*}

\begin{table}[]
\footnotesize
    \centering
    \begin{tabular}{l|l|l|l}
         Label Type & Misinfo $\downarrow$ & Real $\uparrow$ & Effect size \\ \midrule
          Pred & 2.040 & \underline{3.240} & 0.764 \\
          Gold & 2.531 & \underline{3.213} & 1.380 \\ \bottomrule
    \end{tabular}
    \caption{
    Likelihood of news events spreading, i.e. the annotators' rating for how likely it is they would share or read the article based on the shown news event. For ``Pred", we ignore headlines where annotators were unsure about the label. For this and the following tables, arrows indicate the desired direction of the score. We use Cohen's $d$ to compute effect size. } %\ms{mention that effect size is cohen's $d$?}
    \label{table:likelihood} %\ms{numbers are updated}
\end{table}

\begin{table*}[!htb]
    \centering
    \begin{tabularx}{\linewidth}{l|X|l}
         Type & Description & Covered by MRF  \\ \midrule
         Misinformation & Misinformation is an umbrella term for news that is false or misleading. & \LARGE \cmark \\
         Disinformation & Unlike misinformation, disinformation assumes a malicious intent or desire to manipulate. In our framework, we focus on intent in terms of reader-perceived implications rather than questioning whether or not the writer's intentions were malicious given that it is unclear the extent to which original writers might have known article content was misleading. & Potentially  \\
         Fake News & As defined by \cite{10.1257/jep.31.2.211}, fake news refers to \textit{``news articles that are intentionally and verifiably false, and could mislead readers."} \cite{10.1145/3201064.3201100} notes that fake news is a form of hoax, where the content is factually incorrect and the purpose is to mislead. This also overlaps with the definition of disinformation. & Potentially \\
         
         Propaganda & Propaganda is widely held to be news that is \textit{``an expression of opinion or action by individuals or groups, deliberately designed to influence opinions or actions of other individuals or groups with reference to predetermined ends"} \cite{name-calling}. Propaganda is therefore wholly defined in terms of the intent of a writer or group of writers, and may contain factually correct content. & \LARGE \cmark \\
         Satire & We refer to articles written with a humorous or ironic intent as ``satire." We do not explicitly cover satire in MRF, but it is possible that some misinformation articles began as satire and were misconstrued as real news. & Potentially \\
         Real (Trusted) & We consider this to be news that is factually correct with an intent to inform. We note that while real news is distinct from most of the article types shown here, it can also function as propaganda. & \LARGE \cmark \\
    \end{tabularx}
    \caption{Article types based on intention and perceived reliability.}
    \label{table:defs}
\end{table*}

%\subsection{Annotator Statistics}
%\label{sec:demos}

%We provided an optional demographic survey to MTurk workers during annotation. Of the 63 annotators who reported ethnicity, 82.54\% identified as White, 9.52\% as Black/African-American, 6.35\% as Asian/Pacific Islander, and 1.59\% as Hispanic/Latino. For self-identified gender, 59\% were male and 41\% were female. Annotators were generally well-educated, with 74\% reporting having a professional degree, college-level degree or higher. Most annotators were between the ages of 25 and 54 (88\%). We also asked annotators for news preferences. New York Times, CNN, Twitter, Washington Post, NPR, Reddit, Reuters, BBC, YouTube and Facebook were reported as the 10 most common news sources. The main task questions presented to annotators are given in Figure \ref{fig:marsannotation2}.

\subsection{Additional Annotation Details}

We include all annotations from qualified workers in the pilots and final task as part of the dataset, discarding annotations from disqualified workers. We also removed headlines that received no annotations due to deformities in the original text (e.g. unexpected truncation) or vagueness. We paid workers at a rate of \$0.4 per hit during these pilots and \$.6 per hit for the second stage pilot and final task.\footnote{We estimate this to be a fair wage of \$12-\$18/hr, well above minimum wage.} Annotators consent to doing each task by accepting it on the MTurk platform after reading a short description of what they will be asked to do. 

For writer intent implications, we asked annotators if each of the 7 predefined themes was relevant to the event. If a theme was relevant, we asked annotators to provide 1-3 implications related to the chosen theme. For reader perception and action implications, we elicit 1-2 implications. 

All news headlines are in English. The Poynter database contains international news originally presented in multiple languages, however news headlines contained in the database have all been translated into English. 

\subsubsection{Full Instructions to Annotators} 

\begin{quote}
\textit{Thanks for participating in this HIT! You will read a sentence fragment depicting an event from a news article (please note that some of these news articles may contain misinformation). The fragment may contain references to specific organizations or locations. In this case, please write an answer based on the most generic form of this reference (for example if there is a reference to the CDC, provide an answer like ``government," ``government organization" or ``health agency," rather than writing ``CDC."}

\textit{Think about what might be implied by the news event described, including the reaction it might invoke from someone reading about the news event and what the intent of the sentence fragment's author was.}

\textit{The readers' reactions and author's intent may cover multiple topic categories (for example, a sentence fragment may contain implications relating to technology and society), so when thinking about implications try to consider as many topics as possible.}
\end{quote}

\subsection{Post-processing of Annotations} 
\label{sec:postp}

To remove duplicate free-text annotations, we check if annotations along the same MRF dimension have a ROUGE-2 \cite{lin-2004-rouge} overlap of less than .8. If two annotations have a overlap that violates this threshold, we keep one and discard the other. We also remove writer intent annotations that have a ROUGE-2 overlap of greater or equal to .8 with the headline to prevent direct copying. Due to noise in the keyword filtering approach to labeling climate-related NELA-GT headlines, we remove headlines with specific keywords referencing toxic work environments or political climates.\footnote{There may still be cornercases, but this covers the vast majority of mislabelings.} 

Some ``perception" annotations were more suited semantically to being ``action" annotations or vice versa. If an ``action" annotation is a single word categorized as a variant of a \textit{emotion} word \cite{Shaver1987EmotionKF}, we reclassify it as a ``perception." Conversely, if a ``perception" annotation includes ``want," expressing a desire for an action to happen or to do an action, we reclassify it as a ``action." During this process, we also remove single word annotations that feature common misspellings.\footnote{While misspellings were considered during overall quality control of workers, these are difficult to handle automatically. For example, automatic spell-checkers change instances of ``biden" to ``widen," so we forgo automatic spellchecking.} 

We restrict writer intent annotations to be at least three words long. Reader perception and action annotations must be at least three characters in length. 

Finally, we handled missing free-text annotations. If a headline had no free-text annotations, we took this as an indicator of a low-quality example or assumed it lacked enough context for annotators to make a judgement. These invalid headlines were removed (8.5\% of all headlines). If a writer intent annotation is missing, we assume the intent is ambiguous and mark it is as ``unknown intent." These make up 6\% of valid headlines and are not included in the overall count of implications. If a reader perception or action annotation is missing, we infer the corresponding implication from other annotated MRF variables using Table \ref{table:reassign}.  Given the variables in columns 1 and 2, we randomly sample variables from 3 and 4.

%\ec{think some of the details can be moved to appendix}

%\subsection{Topic-level Breakdown}

%\begin{figure}[t]
%    \centering
%    \includegraphics[width=.9\linewidth]{images/Figure_1-7.pdf}
%    \vspace{-1ex}
%    \caption{Breakdown of MBF corpus inferences by topic.}
%     \vspace{-4ex}
%    \label{fig:topic}
%\end{figure}

%We provide a topic-level overview of inferences in Figure \ref{fig:topic}. The distribution of inferences is relatively even across topic categories.

%\subsubsection{Gold Data Evaluation}

%We evaluate aspects of gold inferences to consider potential differences between beliefs held for misinformation headlines compared to real news.  

%\paragraph{Divergence} For articles with more than one unique inference annotated along a particular belief frame dimension, we measure the divergence in beliefs or reactions invoked by the headline by measuring the average cosine distance between pairs of embedded gold inferences. 

%\paragraph{Sentiment} We measure the sentiment of beliefs or reactions invoked by headlines by measuring the \textit{valence} (degree of positivity), \textit{arousal} (degree of emotionality), and \textit{dominance} (degree of agency/control) of lexical content. For this evaluation we use the NRC-VAD lexicon \cite{vad-acl2018}. 

\subsection{Experimental Setup and Model Hyperparameters}
\label{sec:hyper}

All models are trained on either a single Quadro RTX 8000 or TITAN Xp GPU. Average training time for generative models ranges from approx. 1 hour per epoch for T5-base to 4 hours for GPT-2 large. Inference time for models ranges from approx. 10-20 minutes. Average training time for BERT models is approx. 30 minutes per epoch and inference time is approx. 10 minutes. We use a single final training/evaluation run and hyperparameters are manually tuned in the range of 1e-2 to 6e-6. 

\subsubsection{Bert Classification Models}

Supervised classification models are finetuned on our corpus. Both BERT and Covid-BERT models are trained for a maximum of 30 epochs with a learning rate of 1.5e-5 and batch size of 8. Propaganda detection models are trained using the settings given in \cite{da-san-martino-etal-2019-fine}. 
%Examples of predictions made by propaganda detection models  are given in Table \ref{table:examples_ours}, showing that while these models are accurate at predicting rhetorical techniques, such techniques appear in both real and misinformation news headlines.
BERT models have 345M parameters. 

\subsubsection{Generative Models}
For GPT-2, models are finetuned with a learning rate of 2e-5. We use a learning rate of 5e-5 for T5. For all models except GPT-2 large we use a batch size of 16. For GPT-2 large we use a batch size of 4. We use beam search with a beam size of 3 for the generation task. Generation models are trained for a maximum of 10 epochs using early stopping based on dev. loss (in the case of the GPT-2 model finetuned on cancer data we finetune for a single epoch). We optimize using AdamW \cite{Loshchilov2019DecoupledWD} and linear warmup. Model sizes range from 124M parameters for GPT-2 small to 774M parameters for GPT-2 large. 

\begin{figure*}[t]
    \centering
    \includegraphics[width=1\linewidth]{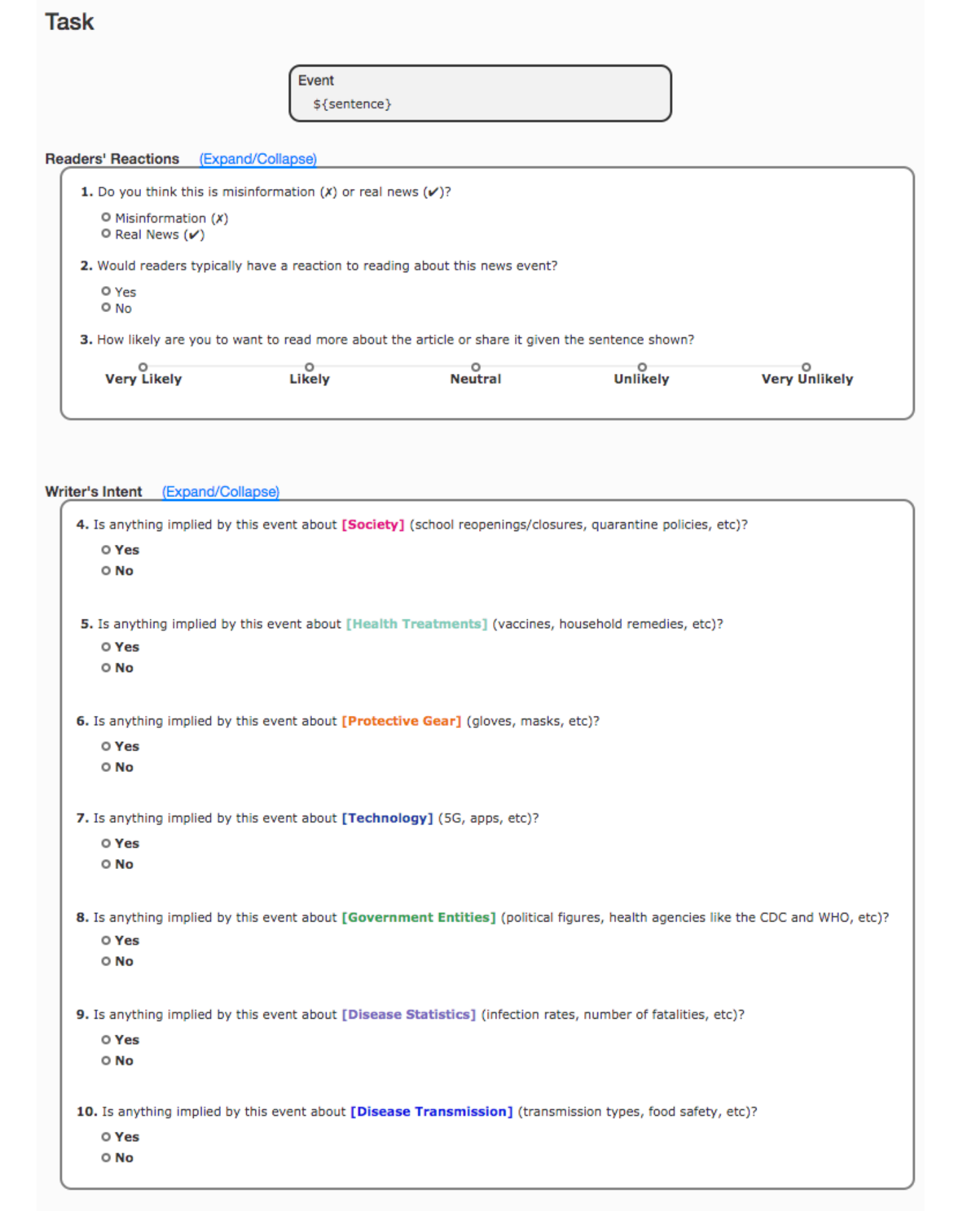}
    \vspace{-4ex}
    \caption{Layout of annotation task for collecting Covid-related MRF data. }
    % \vspace{-4ex}
    \label{fig:marsannotation2}
\end{figure*}

\subsection{Effect of Reader Perception on Article Sharing or Reading} 
\label{sec:reader_effect}

\begin{table*}[!htb]
    \centering
    \small
    \begin{tabularx}{\linewidth}{l|l}
         Headline (Spread) & Pred/Gold \\ \midrule
         Why Companies Are Making Billions of COVID-19 Vaccine Doses That May Not Work (4.0) & Misinfo/Real \\ 
         NATO's Arctic War Exercise Unites Climate Change and WWIII (4.0) & Misinfo/Real\\
          Eat Bugs! EU Pressing member States to Promote Climate Friendly Insect Protein Diets (4.0) & Misinfo/Misinfo \\
          Coronavirus was created in Wuhan lab and released intentionally. (5.0) & Misinfo/Misinfo\\
           \bottomrule
          %Coronavirus was created in Wuhan lab and released intentionally. (5.0) & Misinfo/Misinfo\\
          %COMMENTARY: We Can't Ignore the Harms of Social Distancing (4.0) & Misinfo/Real
          
    \end{tabularx}
    \caption{Headlines that were labeled as misinformation by annotators and also given a high aggregated likelihood of being read or shared (spread). We show the predicted and gold labels.}%\ms{I made the table smaller, feel free to revert}}
    \label{table:examples}
\end{table*}

%\begin{figure}[t]
%
%    \centering
%    \includegraphics[width=1\linewidth,trim={0 0 6cm 0},clip]{images/resize_legend.pdf}
%    \vspace{-4ex}
%    \caption{Distribution of spread (likelihood of sharing) scores in the training set. Aggregated scores are rounded.}
 %    \vspace{-4ex}
%    \label{fig:statsfig}
%\end{figure}%\ec{legend too small}

Annotators tended to be cautious in reported sharing or reading behavior. We found that annotators did have a higher likelihood of sharing or reading real articles over misinformation articles (Table \ref{table:likelihood}), and importantly generally claimed that they would not share or read articles that they thought were misinformation. For 1.2\% of articles reported as misinformation in the training set annotators did provide a likelihood of sharing or reading $\geq 4$. We show examples of articles that were labeled as ``misinfo" but shared or read anyway in Table \ref{table:examples}. While the reasoning for this is unclear, the annotators' reaction frame predictions for reader perceptions and actions may provide insight. For example, annotators were skeptical of the misinformation news event 
``\textit{Coronavirus was created in Wuhan lab and released intentionally.},'' but said they would share/read it anyway and provided ``\textit{readers would feel curious}" and ``\textit{readers would want to know if the wild claim has any truth to it}" as related inferences. Concerningly, this indicates even very obvious misinformation may still be shared or read by generally knowledgeable readers when it contains content they deem particularly interesting or they want to corroborate the article content with others. This aligns with a recent study of 67 million tweets \cite{Huang2020DisinformationAM} that found the “Covid as a bio-weapon started in a lab” theory is a commonly spread disinformation storyline perpetuated by bot-like accounts on Twitter. 

Overall, however, we found that annotators' perception of an article as being more reliable played a positive role in their decision to share or read it. 

\subsection{Analysis of A/B Test} 
\label{sec:more_ab_test}

\begin{table*}[!htb]
    \centering
    \small
    \begin{tabularx}{\linewidth}{X|X|l|l}
         Headline & Generated Writer Intent (Model) & Shift in Trust & Gold Label \\ \midrule
          Every day in Germany more people die because of wrong medical treatment, misuse of drugs or hospital germs than of Covid-19 & The writer is implying that the pandemic isn't that bad (T5-large) & Decreases Trust & Misinfo\\ 
           & & & \\ 
          NYC COVID-19 Deaths During Peak Rivaled 1918 Flu Fatalities & The writer is implying that the pandemic is dangerous (T5-large) & Increases Trust & Real \\
           & & & \\ 
          PCR Tests cannot show the novel coronavirus. & The writer is implying that covid testing is unreliable (GPT-2 large) & Decreases Trust & Misinfo \\
          & & & \\ 
          Alaska's new climate threat: tsunamis linked to melting permafrost & The writer is implying that climate change is real (GPT-2 large) & Increases Trust & Real\\
           & & & \\ 
          "Nearly half of (Missouri) counties have not reported positive (COVID-19) cases." & The writer is implying that covid is not spreading in Missouri (GPT-2 large) & Decreases Trust & Misinfo \\
           & & & \\ 
          Can the catastrophic fires bring some sanity to Australian climate politics? & The writer is implying that wildfires in australia are a result of climate change (GPT-2 large) & Increases Trust &  Real \\
          & & & \\ 
          Wisconsin is ``clearly seeing a decline in COVID infections". & The writer is implying that covid is not spreading in florida (GPT-2 large) & Decreases Trust & Misinfo \\
           \bottomrule
          %Coronavirus was created in Wuhan lab and released intentionally. (5.0) & Misinfo/Misinfo\\
          %COMMENTARY: We Can't Ignore the Harms of Social Distancing (4.0) & Misinfo/Real
          
    \end{tabularx}
    \caption{Examples where generated writer intent implications are effective at changing perceived trustworthiness of news headlines.}
    \label{table:ab_test_examples}
\end{table*}

As shown by Table \ref{table:ab_test_examples}, generated writer intent implications can provide explanations that are effective at increasing reader trust in real news or decreasing trust in misinformation. However, the effect on reader trust is not always indicative of the generated intent's relevance to the headline or accuracy in capturing likely intent. Model errors like hallucinations can also decrease reader trust, as shown in the last example where the wrong state is referenced. This highlights the importance of evaluating effectiveness for both real news and misinformation.

\subsection{Further Related Work}
\label{sec:defs}

 In our framework, we focus on intent in terms of implications rather than questioning whether or not the writer's intentions were malicious given that it is unclear the extent to which original writers might have known article content was misleading. We summarize common definitions for news reliability in Table \ref{table:defs}).
 
%\ms{See my note in Section \ref{sec:dims}. Either we will have discussed all the related work in Section 2, or we have to discuss some more related stuff in a ``further related work'' section?}

%\ms{I think we should cite \cite{schuster2020limitations} somewhere} 

%\ms{Feels like this section should be expanded a little bit... There is a LOT of work on misinfo, so we should at least try to cite some surveys and ack that we only cite a couple of relevant works imo. -- re-visiting this comment, I think we either move the background stuff from section 2 to this section, or we rename this section  ``further related work'' or something so that it's clear that there's more related work elsewhere.}
\paragraph{Rhetorical Framing of News.} Prior work on rhetorical framing \cite[e.g.][]{https://doi.org/10.3732/ajb.0900041, card-etal-2015-media,field-etal-2018-framing} has noted the significant role \textit{media frames} play in shaping public perception of social and political issues, as well as the potential for misleading representations of events in news media. However, past formalisms for rhetorical framing that rely on common writing or propaganda techniques \cite[e.g. \textit{appeal to fear} or \textit{loaded language},][]{da-san-martino-etal-2019-fine} do not explicitly model impact. To that end, we propose a formalism focusing on readers' perception of the writers' intention, rather than specific well-known techniques. %\ec{be specific what are these specific well known techniques}
%\ms{avoid cites in parentheses, instead, use \textbackslash cite[][]\{\}}
\paragraph{Misinformation Detection.} There has been work on integration of knowledge graphs \cite{contentbasedfakenews} and framing detection as a NLI task \cite{Yang2019FakeND}. \citet{zellers2019neuralfakenews} show the effectiveness of using large-scale neural language modeling to detect machine-generated misinformation. Recent work has also highlighted the importance of understanding the impact from misinformation, particularly in health domains \cite{dharawat2020drink,10.1145/3274327}. \citet{10.1145/3395046} and \citet{Hardalov2021ASO} provide comprehensive surveys of misinformation detection methods. Our work is related to stance detection \cite{ghanem-etal-2018-stance}, however our pragmatic frames go beyond understanding the stance of a reader and explicitly capture how reader perceptions affect their actions. 

\paragraph{Countering Misinformation.} It has been noted in prior work that sharing behavior reported in MTurk crowdsourced studies matches behavior in-the-wild \cite{Mosleh2020SelfreportedWT}. \cite{10.1145/3313831.3376213,10.1145/3313831.3376873} show the effectiveness of credibility indicators to persuade readers to decrease their trust in false information. \cite{Jahanbakhsh2021ExploringLI} show that having users assess accuracy of news at sharing time and providing rationales for their decisions decreases likelihood of false information being shared.

\end{document}